\documentclass{article}

\usepackage{PRIMEarxiv}

\usepackage[utf8]{inputenc} 
\usepackage[T1]{fontenc}    
\usepackage{hyperref}       
\usepackage{url}            
\usepackage{booktabs}       
\usepackage{amsfonts}       
\usepackage{amsmath}       
\usepackage{nicefrac}       
\usepackage{microtype}      
\usepackage{lipsum}
\usepackage{fancyhdr}       
\usepackage{graphicx}       
\graphicspath{{media/}}     
\usepackage{cleveref}
\usepackage[table,xcdraw]{xcolor}
\usepackage[inline]{enumitem}
\usepackage{multirow}
\usepackage[normalem]{ulem}
\useunder{\uline}{\ul}{}
\usepackage{pifont}
\usepackage{newunicodechar}
\usepackage{adjustbox}
\usepackage{listings}

\definecolor{bg}{rgb}{0.95,0.95,0.95}
\definecolor{commentcolor}{rgb}{0,0.5,0}
\definecolor{keywordcolor}{rgb}{0.2,0.2,0.7}
\definecolor{stringcolor}{rgb}{0.58,0,0.15}

\lstdefinestyle{mystyle}{
  backgroundcolor=\color{bg},
  basicstyle=\ttfamily\tiny,
  keywordstyle=\color{keywordcolor}\bfseries,
  commentstyle=\color{commentcolor}\itshape,
  stringstyle=\color{stringcolor},
  showstringspaces=false,
  breaklines=true,
  breakatwhitespace=true,
  frame=single,
  tabsize=2,
  numbers=left,
  numberstyle=\tiny\color{gray},
  captionpos=b,
  language=Python
}

\newunicodechar{✓}{\ding{51}}
\newunicodechar{✗}{\ding{55}}

\usepackage[
    style=numeric,
    sorting=none,
    maxbibnames=3,
    giveninits=true
]{biblatex}
\title{Not all tokens are created equal: Perplexity Attention Weighted Networks for AI generated text detection}

\author{
  Pablo Miralles-González, Javier Huertas-Tato, Alejandro Martín, David Camacho \\
  Department of Computer Systems \\
  Technical University of Madrid \\
  Madrid\\
  \texttt{\{pablo.miralles, javier.huertas.tato, alejandro.martin, david.camacho\}@upm.es} \\
}

\begin{document}
\maketitle

\begin{abstract}
The rapid advancement in large language models (LLMs) has significantly enhanced their ability to generate coherent and contextually relevant text, raising concerns about the misuse of AI-generated content and making it critical to detect it. However, the task remains challenging, particularly in unseen domains or with unfamiliar LLMs. Leveraging LLM next-token distribution outputs offers a theoretically appealing approach for detection, as they encapsulate insights from the models' extensive pre-training on diverse corpora. Despite its promise, zero-shot methods that attempt to operationalize these outputs have met with limited success. We hypothesize that one of the problems is that they use the mean to aggregate next-token distribution metrics across tokens, when some tokens are naturally easier or harder to predict and should be weighted differently. Based on this idea, we propose the Perplexity Attention Weighted Network (PAWN), which uses the last hidden states of the LLM and positions to weight the sum of a series of features based on metrics from the next-token distribution across the sequence length. Although not zero-shot, our method allows us to cache the last hidden states and next-token distribution metrics on disk, greatly reducing the training resource requirements. PAWN shows competitive and even better performance in-distribution than the strongest baselines (fine-tuned LMs) with a fraction of their trainable parameters. Our model also generalizes better to unseen domains and source models, with smaller variability in the decision boundary across distribution shifts. It is also more robust to adversarial attacks, and if the backbone has multilingual capabilities, it presents decent generalization to languages not seen during supervised training, with LLaMA3-1B reaching a mean macro-averaged F1 score of 81.46\% in cross-validation with nine languages.
\end{abstract}

\keywords{AI Generated Text Detection \and Perplexity-based methods}

\section{Introduction}\label{sec:introduction}

The proliferation of large language models (LLMs) has ushered in a new era of text generation, where artificial intelligence can produce coherent, contextually relevant content with remarkable fluency. These advancements hold transformative potential across industries, yet they also raise serious concerns about the misuse of AI-generated text~\cite{solaiman2019ReleaseStrategies}. The ability to massively produce misinformation or automatically solve school assignments underscores the urgent need for effective detection mechanisms. However, identifying AI-generated text is far from straightforward, particularly in scenarios involving unseen domains or unfamiliar generative models~\cite{li2024MAGEMachinegenerateda,wang-etal-2024-m4}, or adversarial attacks~\cite{dugan2024RAIDShareda,li2024MAGEMachinegenerateda}. The diversity in writing styles, prompt configurations, and the opaque nature of AI systems contribute to the complexity of this task.  

Recent research has explored various strategies for detecting AI-generated text, with approaches based on next-token distribution output emerging as promising (e.g. Binoculars~\cite{hans2024SpottingLLMs} or DetectLLM~\cite{su2023DetectLLMLeveraging}). These outputs, which represent likelihood estimates calculated by LLMs during text generation, encapsulate knowledge derived from extensive pre-training on diverse corpora. They offer a theoretically grounded foundation for detection, leveraging the statistical properties of language encoded in LLMs. Despite their appeal, zero-shot methods that directly utilize these next-token distribution outputs have shown limited effectiveness and performance (for example, DetectGPT~\cite{mitchell2023DetectGPTZeroShot} results in MAGE~\cite{li2024MAGEMachinegenerateda}).

One of the limitations of current zero-shot methods is their reliance on simple aggregation techniques, such as averaging next-token metrics across all tokens. This uniform treatment of tokens disregards the intrinsic differences in predictive complexity. For instance, predicting the completion of a word is often straightforward, while initiating a sentence, with its broader range of possible continuations, is inherently more challenging. The beginning of a text, unconditioned due to the unavailability of the generating prompt, is also naturally more random. Recognizing these nuances is critical to improving detection performance.  

To address this gap, we propose the Perplexity Attention Weighted Network (PAWN), a novel approach that assigns dynamic weights to tokens based on their semantic, contextual and positional significance. PAWN leverages the semantic information encoded in the last hidden states of the LLM and positional indices to modulate the contribution of each token to a specific feature based on the next-token distribution metrics. Our method refines the aggregation process, resulting in more accurate and robust detection. Unlike zero-shot approaches, PAWN involves lightweight training, but it mitigates resource constraints by enabling the caching of hidden states and next-token distribution metrics on disk.  

Empirical evaluations demonstrate that PAWN achieves competitive and even superior performance compared the best baselines (fine-tuned encoder LMs). Notably, our model generalizes better to unseen domains and generative models, exhibiting consistent decision boundaries across distribution shifts and enhanced resilience to adversarial attacks. Furthermore, if the backbone LLM presents multilingual capabilities, PAWN showcases decent generalization to languages not seen during supervised training. These results highlight the potential of PAWN as a practical and resource-efficient solution for detecting AI-generated text.

The main contributions of this work are summarized as follows:
\begin{itemize}
    \item We propose a novel detection framework, \textbf{Perplexity Attention Weighted Network (PAWN)}, which dynamically weights next-token distribution metrics based on semantic information from the LLM's hidden states and positional information. This model:

    \begin{itemize}
        \item Despite not being zero-shot and requiring supervised training, has a \emph{very small number of training parameters} (in the order of one million). The backbone is thus frozen, allowing us to cache the hidden states and metrics on disk, significantly reducing resource requirements compared to fine-tuning large models.
        
        \item Demonstrated \emph{competitive or superior performance} to the strongest baseline (fine-tuned encoder LMs) in in-distribution detection.
        
        \item Showcased better generalization capabilities in \emph{unseen domains and with unfamiliar generative models}, exhibiting more stable decision boundaries.
        
        \item Although still vulnerable to paraphrasing attacks, it achieved \emph{better overall robustness to adversarial attacks}, mitigating vulnerabilities in AI-generated text detection.
        
        \item If the backbone model has multilingual capabilities, \emph{PAWN achieves decent performance in languages not seen during supervised training}.
    \end{itemize}
    
    \item By performing ablation studies with the different branches of the PAWN model, we study the effect of using semantic information from hidden states and next-token distribution metrics. As one might expect, we find that the former results in better performance in-distribution but worse generalization. By limiting the role of semantic information to weighting the metric-based features, PAWN achieves both great in-distribution performance and good generalization.
\end{itemize}

\section{Related work}\label{sec:sota}
This section reviews the state-of-the-art in the field of AI-generated text detection, highlighting key advances and challenges. We begin the discussion with the latter. These challenges have shaped how the most recent and advanced datasets have been constructed, which we discuss next. We end the section with a classification of the main detection methods found in the literature.

\subsection{Challenges in detecting AI-generated text}
The detection of AI-generated text has been shown to face numerous challenges, preventing current detectors from being used reliably in real world scenarios. Some of these challenges are the following.

\begin{enumerate}[label=(\roman*)]
    \item \textbf{Generalization to unseen domains and models.} It has been shown now in several works~\cite{li2024MAGEMachinegenerateda,wang-etal-2024-m4} that trained detectors often suffer when applied to a domain or source generative model that is not seen during training. This makes it very difficult to reliably apply the detectors in real-world scenarios, as in most cases we do not know the domain or source model beforehand.
    
    This is especially challenging in the current \textbf{continuously changing landscape of generative models}, where new and improved LLMs are being developed and made available to the public at an incredible rate.
    
    \item \textbf{Multilingual detection.} The problem of generalizing to new domains goes one step further in multilingual detection. This is especially true for low-resource languages, where supervised data is scarce, and so is pre-training data for many of the backbone LMs used in detectors. These LMs also suffer from tokenization problems, as many tokens are required to encode texts in rare languages, especially in other alphabets.
    
    These problems were illustrated in~\cite{wang-etal-2024-m4}, where \citeauthor{wang-etal-2024-m4} found a great degradation in performance in languages where the model was not fine-tuned, even with a multilingual backbone model such as XLM-RoBERTa~\cite{conneau2020UnsupervisedCrosslingual}.
    
    \item \textbf{Adversarial attacks.} Users of generative models are constantly devising strategies to avoid detection. One common strategy is the use of paraphrasing attacks~\cite{kirchenbauerWatermarkLarge,krishna2023paraphrasing}, where the generated text is automatically processed by the same model or another to paraphrase sentences in the text. Although simple, these attacks have been shown to greatly degrade performance~\cite{li2024MAGEMachinegenerateda}. Other very simple attacks are available~\cite{dugan2024RAIDShareda}, such as the insertion of invisible characters, whitespace and homoglyphs, or the use of intentional misspelling.
    
    
    \item \textbf{Interleaved human and machine-generated text.} Many times, texts are products of human and machine collaboration. The users might create an initial template to work with and then rewrite some part according to their need. This opens the way to a different and related problem: detecting AI-generated spans within a text~\cite{wang2024M4GTBenchEvaluation}. We do not treat this problem in this work.
\end{enumerate}

\subsection{Datasets and benchmarks}

The foundation for effective AI-generated text detection lies in robust datasets and evaluation benchmarks. Although there is a wealth of work in this research line, it was not until very recent work that evaluation sets started to address the challenges detectors faced when applied in real-world scenarios. For example, the data from the Voight-Kampff Generative AI Authorship Verification Task at PAN and ELOQUENT 2024~\cite{bevendorff:2024} and the TuringBench benchmark~\cite{uchendu2021TURINGBENCHBenchmark} do not include information on the domain. The Ghostbuster dataset~\cite{verma2024GhostbusterDetecting}, HC3~\cite{guo2023HowClose} and the GPT2 output dataset~\cite{openai2024openaigpt2outputdataset} only include texts from a single source model. In this work, we focus on three modern datasets that address generalization to both new domains and new source models, and that contain a large number of examples.

\textbf{MAGE~\cite{li2024MAGEMachinegenerateda}} contains almost half a million examples, with a focus on quality and diversity. At the time of writing, MAGE contains the most diverse set of domains and source generative models. In addition, they standardize a set of eight testbeds to evaluate the detectors in different settings and run several existing baselines on them. A \textbf{main corpus} of data is provided, where human- and machine-generated texts are sourced from ten different domains. A \textbf{separate corpus} containing samples from a complementary set of four domains is given as well. The \textbf{main corpus} sources machine-generated texts from 27 generative models, grouped into seven families. The \textbf{separate corpus} contains texts generated by GPT-4, a model that is not included in the main corpus. The authors also include a set with adversarially attacked texts by applying a paraphrasing model sentence-by-sentence.

\textbf{M4~\cite{wang-etal-2024-m4,wang2024M4GTBenchEvaluation}} is a slightly less diverse and smaller dataset of $122K$ examples. They provide two separate sets, one with English texts only and one with texts from multiple languages. We find texts sourced from six different English domains, a smaller set than that of MAGE. However, they include more domains from eight extra languages:
\begin{enumerate*}[label=(\roman*)]
    \item Arabic,
    \item Bulgarian,
    \item Chinese,
    \item Indonesian,
    \item Russian,
    \item German,
    \item Italia,
    \item and Urdu.
\end{enumerate*}
The number of models included is also much smaller than in MAGE, with the monolingual set containing samples from six different models. The multilingual set includes samples from two additional models in the Italian and Arabic languages.

\textbf{RAID~\cite{dugan2024RAIDShareda}} is a massive dataset of $6$ million examples from a wide range of domains and source generative models. The variety of selected models is somewhere between M4 and MAGE, as they include texts sourced from eleven different models. The novelty of RAID is that they also use two different decoding strategies, greedy and sampling, as well as repetition penalty. These features are included in the dataset, and we use them to test the generalization of our model in new decoding settings. As for domains, RAID sources from eight domains, including factual knowledge, generalization and reasoning, creative and conversational abilities, and familiarity with specific media such as books and reviews.

The authors provide examples attacked with a variety of adversarial strategies: replacing words synonyms; deleting articles such as “the,” “a,” and “an”; converting to British spelling; inserting paragraph breaks between sentences; swapping uppercase and lowercase letters; replacing characters with visually similar Unicode homoglyphs; shuffling digits within numbers; rephrasing text using a fine-tuned T5 model; introducing common misspellings; adding extra spaces between characters; and inserting zero-width spaces between every other character.

Unfortunately, the authors do not include results from baselines trained in the RAID training split. Instead, their baselines were trained on a different dataset each. The authors did not provide a testbed selection either.

\subsection{Detection methods}

There is a large and growing body of work focused on detection methods for AI-generated text. We review some of the main methods that appear in the literature, breaking them into four categories. The first category includes watermarking methods, which add hidden patterns to AI-generated text to make it identifiable. Zero-shot detectors rely on large language models (LLMs) pre-trained for next-token prediction to identify AI-generated content without additional training, using the next-token distributions of the text to generate statistics that separate human- and machine-generated examples. Other approaches involve fine-tuning language models and training them specifically for the detection classification task. Finally, LLMs have also been prompted to detect whether a text is AI-generated, asking them to complete the given prompt with the answer.

\subsubsection{Watermarking technology}
Watermarking technologies offer a proactive approach to text detection by embedding identifiable markers in AI-generated outputs. These methods are not strictly comparable to other types of detector, as they require the cooperation of the LLM provider.

An example of these methods is given by \citeauthor{kirchenbauerWatermarkLarge}~\cite{kirchenbauerWatermarkLarge}. They proposed an algorithm that uses the hash of the last token and a random number generator to produce a list of red tokens that are banned or made unlikely to be selected. Thus, a text that has been generated with this algorithm can be detected because most of the tokens will be in the green list, that is, not in the red list. However, human text is expected to result in half of the tokens in each list. To avoid reducing the quality of the generated text, they applied a positive bias to the logits of tokens in the green list. In this way, sampling from very sharp distributions with low entropy is not greatly altered, and only high-entropy distributions with many available options as the next token are substantially modified.

The authors found that this method can be effectively attacked by using another LLM to paraphrase spans of the text. In fact, the robustness of the method to such attacks was studied by \citeauthor{kirchenbauer2024on} in~\cite{kirchenbauer2024on}. Other attacks have been pointed out in the literature. A simple example is to ask the LLM to produce an emoji after each generated token, removing them from the sequence afterwards.

\subsubsection{Zero-shot detectors}\label{sssec:zero-shot}

Many zero-shot detection methods have been proposed in prior work. These methods are attractive because they do not require fine-tuning of large language models, but only selecting a boundary between machine-generated and human-generated texts for some score or metric they produce. We review some of the main methods in this category, all of them employing the next-token distribution of a frozen LLM to calculate some metric for the text.

\citeauthor{gehrmann2019GLTRStatisticala}~\cite{gehrmann2019GLTRStatisticala} proposed GLTR. They used the probability of the token that occurred next, the rank of that token, and the entropy of the next-token distribution to create a visualization tool for human detectors, showing different colors for each token depending on these metrics.

\citeauthor{mitchell2023DetectGPTZeroShot}~\cite{mitchell2023DetectGPTZeroShot} proposed DetectGPT. They applied local perturbations to the text using a different model such as T5~\cite{JMLR:v21:20-074}. They mask some of the tokens and apply the mask-filling model to create perturbed versions of the text. Then, they use the average log-probability of both the original text and the sampled perturbed versions, subtracting the former with the mean of the latter.

DetectGPT is very computationally expensive as it runs the paraphrasing model and detection LLM once per perturbed sample. To solve this, \citeauthor{bao2024fastdetectgpt}~\cite{bao2024fastdetectgpt} presented FastDetectGPT. Their approach re-samples tokens independently of each other, using the same LLM as the one used for detection. In other words, after resampling one token, they do not regenerate the text that follows but keep all the following next-token distributions the same for sampling. This allows them to avoid multiple runs through the model. Despite the increased efficiency, they also seemed to increase the detection performance of the model.

\citeauthor{su2023DetectLLMLeveraging}~\cite{su2023DetectLLMLeveraging} proposed two methods, called DetectLLM-LRR and DetectLLM-NPR. The former leverages the ratio between the next-token log-probability and log-rank. The latter compares the average log-rank information of the text with the mean of the same metric across a range of perturbed versions of the text.

\citeauthor{hans2024SpottingLLMs}~\cite{hans2024SpottingLLMs} proposed Binoculars. In their method, they use two very similar models (they must at least share the same tokenizer) called the observer and the performer, often the foundation and instruction-tuned versions of the same model. First, they used the observer model to calculate the average log-probability of the text. Next, they compute the average cross-entropy between the next-token distributions of the observer and performer models, which measures how surprising the output of one model is to the other. The final score is the ratio of these two metrics. This normalization is motivated by what they call the ``capybara problem''. Without the prompt available to condition the text, the models will naturally assign higher perplexities depending on factors such as topic or style.

Despite being very interesting ideas and not requiring supervised training, these detectors are often not the best performing ones (see e.g.~\cite{li2024MAGEMachinegenerateda,wang2024M4GTBenchEvaluation}), falling behind fine-tuned LMs.

\subsubsection{Fine-tuned LMs}

As in any other text classification problems, fine-tuning pre-trained language models (LMs) has emerged as a powerful alternative. Encoder-only pre-trained models such as BERT~\cite{devlin2019BERTPretraining}, RoBERTa~\cite{liu2019RoBERTaRobustly} or XLM-RoBERTa~\cite{conneau2020UnsupervisedCrosslingual} appear in a large body of work, with OpenAI~\cite{solaiman2019ReleaseStrategies} even releasing their own RoBERTa large fine-tuned detector. Although simple, these methods very often provide the strongest detection baselines~\cite{li2024MAGEMachinegenerateda,wang2024M4GTBenchEvaluation}. Unfortunately, they also tend to suffer greatly in out-of-distribution domains and generative models~\cite{li2024MAGEMachinegenerateda,wang2024M4GTBenchEvaluation}.

\citeauthor{hu2023radar}~\cite{hu2023radar} fine-tune a RoBERTa-large model using the RADAR framework, adversarially training against a paraphrasing LLM that learns with reinforcement learning (PPO) to beat the classifier. While it doesn't improve general detection performance, it does increase robustness to paraphrasing attacks greatly.

\citeauthor{liu2024DoesDetectGPT}~\cite{liu2024DoesDetectGPT} proposed a different idea. They used perturbed examples as DetectGPT, but instead of using LLM metrics, they trained an encoder model with contrastive learning, using masked and filled versions of the text as anchors. They also employed the YAKE algorithm~\cite{campos2020YAKEKeyword} to weight the importance of each token, using this information to prioritize masking those tokens and to weight the pooling layer of the encoder.

Another example of contrastive learning is given by \citeauthor{guo2024detective}~\cite{guo2024detective}, training an encoder with texts being considered positive examples if they were generated by the same LLM or by humans. This contrastive learning objective is used as an auxiliary task to the classification one, training simultaneously with the idea of improving generalization.

\subsubsection{LLMs as detectors}
Large Language Models (LLMs) themselves can serve as detectors for AI-generated text, using their advanced understanding of linguistic patterns. In particular, we refer to methods that include the text in the prompt and ask the LLM whether it is machine-generated or human-generated.
For example, \citeauthor{bhattacharjee2024FightingFire}~\cite{bhattacharjee2024FightingFire} used ChatGPT as detector, although it did not prove to be reliable. \citeauthor{koike2024OUTFOXLLMGenerated}~\cite{koike2024OUTFOXLLMGenerated} further extend on this idea by using in-context learning (ICL), providing semantically similar labeled examples in the prompt. This methodology, together with their framework to produce adversarial examples, seems to improve the performance of the methods.

\citeauthor{li2025LearningRewrite}~\cite{li2025LearningRewrite} proposed an interesting approach. They argue that LLMs tend to modify AI-generated text less than human text when asked to paraphrase. Based on this, they further fine-tune LLMs to make fewer changes to AI-generated texts and more to human ones. While they show promising results, they do not compare themselves with strong baselines or in the newest and hardest benchmarks.


\section{Methodology}\label{sec:method}

\subsection{Perplexity Attention Weighted Network (PAWN)}\label{ssec:model}

Like the zero-shot methods we reviewed in \cref{sssec:zero-shot}, we use pretrained LLMs to obtain the next-token distributions from the given text. From these distributions, we generate five different metrics, measuring the likelihood of the next token and the randomness of the distribution. 

However, zero-shot methods often use a simple mean to aggregate different metrics.
Our main addition is the use of semantic and positional information to weigh the relevance of each token. The reason for this is that some of the tokens are naturally easier or harder to predict. Consider two examples. First, completions of words are expected to be very easy for the LLM, as the possibilities are very reduced, and patterns are not too difficult. Second, the beginning of a text is always very unconditioned, making the first few tokens naturally very random. Thus, applying a simple aggregation where every token is equally weighted will add a lot of unnecessary noise. To address this, we used both semantic information from the last hidden states of the LLM and positional information to perform some filtering.

\begin{figure}[t]
    \centering
    \includegraphics[width=1\linewidth]{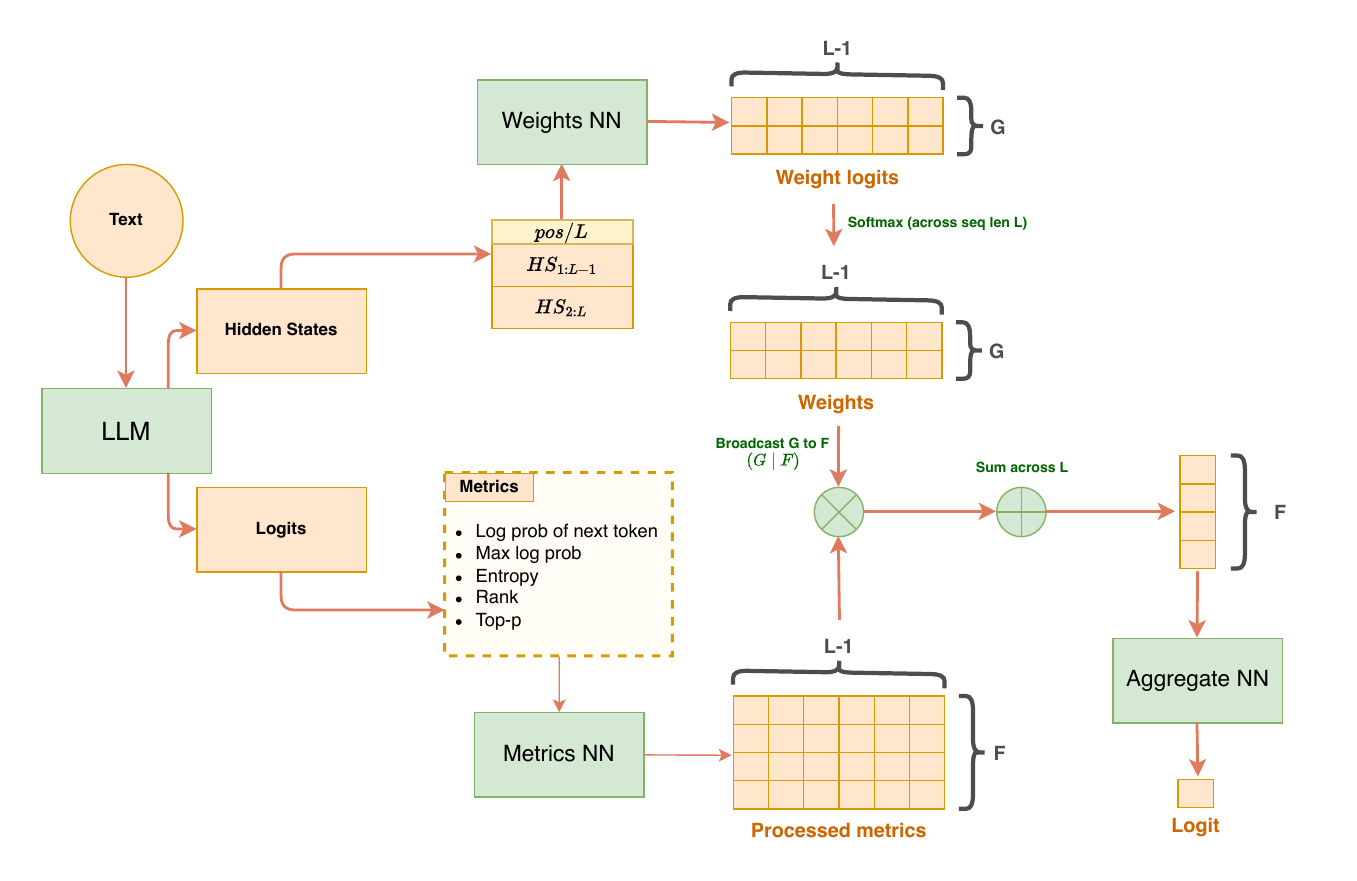}
    \caption[Diagram of the Perplexity Attention Weighted Network]{Diagram of the Perplexity Attention Weighted Network.}
    \label{fig:pawn-diagram}
\end{figure}

\Cref{fig:pawn-diagram} shows the full diagram of our model. We explain it in four steps, both the computations and the rationale behind each design decision.

\begin{description}
    \item[LLM run.] The texts initially go through a selected decoder-only LLM pre-trained for next-token prediction. This returns both the last hidden states and the logits of the next-token distribution.
    
    In this work, we use \texttt{openai-community/\allowbreak  gpt2}~\cite{radford2019language} and \texttt{meta-llama/\allowbreak Llama-3.2-1B-Instruct}~\cite{grattafiori2024Llama3}, covering small and medium model sizes.
    
    \item[Computing metric features.] From the output logits we generate five metrics that measure different aspects of the distribution, stated individually in the enumeration below, instead of using only the probability or log-probability of the next token. These metrics are processed by an MLP network to generate \(F\) different \emph{metric features}.    
    
    The five metrics are listed below. They are also mathematically formulated for a sequence of tokens \(t_1, \dots, t_N\) of length $N$, and an LLM next-token distribution probability matrix \(P \in \mathcal{M} ^{N \times V}\), with $V$ being the size of the vocabulary. An intuition of their relevance in detecting AI-generated text is given after the enumeration.
    
    \begin{enumerate}[label=(\roman*)]
        \item \textbf{Log-probability of the token that occurred next in the sequence.} This metric is formulated as
        \[
        M^{\text{log-prob}}_i = \log P_{i,t_{i+1}}
        .\]
        
        \item \textbf{Entropy of the distribution.} This metric measures the randomness in the distribution, that is, how spread the probability mass is across many tokens. In mathematical terms it is expressed as
        \[
        M^{\text{entropy}}_i = \sum_{j=1}^V P_{i,j} \log P_{i,j}
        .\]
        
        \item \textbf{Maximum log-probability.} Another measure of randomness, useful to compare with the log-probability of the token that occurred. If they are close, then the token was one of the most probable ones. This can be interesting in distributions where the mass is spread across a few tokens. For example, in the phrase ``\verb|My favorite color is|'' the probability is distributed among the different colors. Mathematically:
        \[
        M^{\text{max-log-prob}}_i = \max_{j=1,\dots, V} \log P_{i,j}
        .\]
        
        \item \textbf{Rank of the token that occurred next in the sequence.} The quantile that the token that occurred next represented in the next-token distribution. In other words, the minimum \(K\) parameter in Top-K decoding~\cite{holtzman2020curiouscase} where the token could have been selected, divided by the size of the vocabulary. It can be expressed mathematically as
        \[
        M^{\text{rank}}_i = \mathrm{rank}\left( \log P_{i,:}, t_{i+1} \right) / V
        .\]
        
        \item \textbf{Sum of probabilities of tokens with higher or equal probability than the token that occurred}. In other words, the minimum \(P\) parameter in Top-P decoding~\cite{holtzman2020curiouscase} where the actual token is covered with its full probability. It is formulated as follows:
        \[
        M^{\text{top-p}}_i = \sum_{j=1,\dots, V; P_{i,j} \geq  P_{i,t_{i+1}}} P_{i,j}
        .\]
        
    \end{enumerate}
    
    All of these extra metrics serve one of two purposes. The first is to get a metric that is closer to the Top-P or Top-K decoding algorithms~\cite{holtzman2020curiouscase}, which are often used during text generation. These metrics can help in assessing if a token could have been produced by using that decoding strategy. The second is measuring the randomness of the distribution. This helps in contextualizing the log-probability of the token, as it is not the same to have a low probability in a very random distribution where the probability mass is very distributed, than in a very sharp distribution with a single very likely token. But it is also useful as a metric itself: we expect human text to take the LLM to uncharted territory more often.

    \item[Computing weights.] Next, the hidden states of the LLM and positional information are used to filter out and prioritize the different tokens. Instead of just using the corresponding hidden state, we also concatenate the hidden state of the next token. Both of these tokens are relevant when deciding which metrics to take into account and which to discard, and we found a slightly higher performance empirically.
    
    As positional information, we append the index of each token, divided by a fixed maximum length to avoid very large numbers. Again, the rationale behind it is that the first tokens of the sequence are very unconditioned and will likely have a low log-probability, so we want the model to be able to take this into account.
    
    All of these features are processed by an MLP network, generating \(G\) \emph{feature logits}, where \(G\) divides \(F\). These new \emph{feature logits} are converted to \emph{feature weights} by applying a softmax operation across the length of the sequence. 

    \item[Aggregating metrics.] The \(G\) \emph{feature weights} are broadcasted up to the dimension \(F\). They are then used to compute a weighted sum of the \emph{metric features} across the length of the sequence, summarizing the full text into a single feature vector.

    By using the softmax function, we restrict the role of the hidden states to weighting, and prevent hidden state and positional information from appearing in the summary vector, at least directly. This is because any component of the \emph{metric features} that is independent of the metrics, such as bias parameters, will be summed up to itself after being aggregated with weights summing up to one.
    
    \item[Processing summary vector.] The summary vector goes through a final MLP network to produce the logit or probability of the text being generated by AI.
    
\end{description}

To further illustrate the method and the intuition behind it, we provide PyTorch pseudocode in \Cref{sec:pseudocode}. We also visualize the effect of the token's position on the resulting token weights, and the most and least weighted tokens on average across several examples in \Cref{sec:qualitative analysis}.

\subsection{Data}

As discussed in~\cref{sec:sota}, our choice of datasets was mainly motivated by whether they addressed the challenges detectors face when applied in the real world. Thus, we select MAGE~\cite{li2024MAGEMachinegenerateda}, M4~\cite{wang-etal-2024-m4,wang2024M4GTBenchEvaluation} and RAID~\cite{dugan2024RAIDShareda}, as they include texts from a variety of domains and source generative models. We also find different languages, decoding hyperparameters and adversarial attacks.

\subsection{Technical details}
\subsubsection{Hyperparameters}

In \Cref{tab:hyperparams} we show the hyperparameters we used for the PAWN model with both backbone models \texttt{openai-community/\allowbreak gpt2}~\cite{radford2019language} and \texttt{meta-llama/\allowbreak Llama-3.2-1B-Instruct}~\cite{grattafiori2024Llama3}. The MLP hyperparameters are common to all MLP networks in PAWN, that is, the networks are similar except for the input and output dimensions. We note that we drop out \(15\%\) of tokens during training by setting their \emph{gate logits} to \(-\infty\) as a form of regularization.

\begin{table}[!ht]
\centering
\begin{adjustbox}{max width=\textwidth}
\begin{tabular}{c|cccc|cccc}
\toprule
LLM  & \multicolumn{4}{c|}{PAWN}                        & \multicolumn{4}{c}{MLP}                                 \\
Max tokens      & Pos. encoding & \# gates G & \# features F & Dropout tokens & \# hidden layers & \# hidden feat. & Norm. & Act. \\
\midrule
512             & Indices       & 256        & 256           & 0.15           & 3                & 256                 & Layer         & GELU      \\
\bottomrule
\end{tabular}

\end{adjustbox}
\caption{Hyperparameters used to train the Perplexity Attention Weighted Network (PAWN). MLP hyperparameters are common to all three MLP networks in PAWN. Dropout tokens is the ratio of tokens that are masked during training by setting their \emph{gate logit} to \(-\infty\).}
\label{tab:hyperparams}
\end{table}

\subsubsection{Parameter counts}

One question that arises is the fairness of the comparisons with fine-tuned LM baselines given the size of the backbone model. \Cref{tab:nparams-mage} shows the parameter count of the main LM baselines and of the PAWN models we used, separating the backbone and head counts. We find that the PAWN model with a \texttt{openai-community/gpt2} backbone has the second fewest total parameters. The number of trainable parameters is two orders of magnitude lower than that of the fine-tuned LMs, regardless of the backbone. However, the \texttt{meta-llama/Llama-3.2-1B-Instruct} backbone greatly increases the number of total parameters, and this should be taken into account when comparing the results.

\begin{table}[!ht]
\centering
\begin{adjustbox}{max width=0.6\textwidth}
\begin{tabular}{r|c}
\toprule
Module                           & Number of parameters \\
\midrule
FacebookAI/roberta-base          & 125M                 \\
FacebookAI/roberta-large         & 355M                 \\
FacebookAI/xlm-roberta-base      & 279M                 \\
Longformer                       & 149M                 \\
openai-community/gpt2            & 137M                 \\
meta-llama/Llama-3.2-1B-Instruct & 1.24B                \\
\midrule
PAWN-GPT2 head                   & 989K                 \\
PAWN-LLaMA head                  & 1.6M                 \\
\bottomrule
\end{tabular}
\end{adjustbox}
\caption{Parameter counts of the LM baselines and backbones used in this work, together with those of the PAWN networks with backbones \texttt{openai-community/gpt2} \texttt{meta-llama/Llama-3.2-1B-Instruct}.}
\label{tab:nparams-mage}
\end{table}

\subsubsection{Inference efficiency}

To supplement the discussion about the number of parameters, we study the  inference efficiency of PAWN with our two selected backbones and of the main baselines used in this work. This is measured in terms of runtime per batch and peak memory usage for batch sizes 1 and 16, benchmarking non-batched and batched inference.

In terms of inference time, the comparison generally matches what we expect from parameter counts. The only notable exception is Longformer, seemingly slower than other models in its parameter class. Surprisingly, the average inference time of PAWN-LLaMA is not too far behind that of RoBERTa-large, despite being three times larger. PAWN models seem to suffer more in terms of peak memory consumption, especially with large batches. The reason is that we have to compute and operate on log probabilities with a very large vocabulary, resulting in several massive matrices. This can be optimized during inference by processing the logit matrix of each element in the batch separately, reducing the peak memory usage to be closer to the 1-element batch value, which is still larger than for encoder-only models with the same parameter counts. We did not implement this optimization. The slowest and heaviest detector by far is Binoculars, which is to be expected given that it leverages two 7B LLMs. Its peak memory does not spike as heavily during batched inference as PAWN-LLaMA. This is because the vocabulary of \texttt{meta-llama/Llama-3.2-1B-Instruct} is twice as big as the vocabulary of \texttt{tiiuae/falcon-7b}, and because they perform fewer operations on the massive logit matrix.

\begin{table}[!ht]
\centering
\begin{adjustbox}{max width=\textwidth}
\begin{tabular}{r|cc|cc}
\toprule
              & \multicolumn{2}{c|}{1 sample / batch} & \multicolumn{2}{c}{16 samples / batch} \\
              & Batch time (s)   & Peak memory (GB)  & Batch time (s)    & Peak memory (GB)   \\
\midrule
PAWN GPT2     & 0.0090±0.0274    & 0.61              & 0.1144±0.1311     & 5.77               \\
PAWN LLaMA-1b & 0.0227±0.0526    & 3.33              & 0.4853±0.1815     & 18.48              \\
RoBERTa-base  & 0.0066±0.0134    & 0.49              & 0.1077±0.0616     & 0.75               \\
RoBERTa-large & 0.0153±0.0182    & 1.35              & 0.3419±0.0409     & 1.69               \\
Longformer    & 0.0377±0.0132    & 0.63              & 0.2390±0.0618     & 1.61               \\
Binoculars    & 0.1902±0.0965    & 26.43             & 3.9532±0.2004     & 32.84              \\
\bottomrule
\end{tabular}
\end{adjustbox}
\caption{Efficiency of different models during inference. We used a total of 4096 English samples with a mean text length of 1240 characters, with models truncating at 512 tokens. We show results for batched and non-batched inference. The SDPA attention implementation from PyTorch is used for all models. Batch time is reported as \emph{mean ± std}. Binoculars is implemented with the \texttt{tiiuae/falcon-7b-instruct} and \texttt{tiiuae/falcon-7b} models.}
\label{tab:efficiency-benchmark}
\end{table}

\subsubsection{Caching}
Even though our model requires supervised training, the backbone LLMs are frozen, greatly reducing training compute requirements. Further, we cache the last hidden states and metrics for each of the data samples in MAGE, reducing training time by a factor between 7 and 10 with our hardware and model selection. Thus, only one iteration of the LLM in inference mode over the dataset is required for training, massively reducing training resource requirements.

\subsubsection{Setup}

All experiments were run on a single NVIDIA GeForce RTX 3090 GPU with 24GB of memory, except for Binoculars which required two such GPUs. The source code is available at \url{https://github.com/pablomiralles22/ai-gen-detection}, together with the Conda environment file containing the library versions we use.
\section{Experiments and results}\label{sec:results}

\subsection{In-distribution performance and generalization to unseen domains and models on MAGE}

Our main experiments are performed on the MAGE dataset~\cite{li2024MAGEMachinegenerateda}. The authors provided a fixed set of settings to evaluate the detectors, starting with six testbeds from the \textbf{main corpus}. In the first four, detectors are tested in-distribution, that is, on the same domains and source models they were trained on. The four settings they provide include training and testing on \emph{one domain \& one source model (TB1)}, on \emph{one source model \& all domains (TB2)}, on \emph{all source models \& one domain (TB3)} and on \emph{all source models \& all domains (TB4)}.

Next, we find two testbeds that evaluate the detectors out-of-distribution, that is, on unseen domains or source models. To do this, they proposed leave-one-out experiments across \emph{domains (TB5)} and \emph{source models (TB6)}. They create splits dividing the samples by domain and model family, respectively. For each set, a model is trained on the samples that do not belong and tested on the samples that do. The results are averaged across sets.

Finally, the authors of MAGE generate two additional testbeds from the \textbf{separate corpus}. These testbeds go one step further in evaluating the detector's ability to generalize out-of-distribution. The first one includes an \emph{unseen domain and source model (TB7)}, evaluating models trained on the main corpus in a different one. The second one includes \emph{paraphrasing attacks (TB8)}, testing the detector's robustness to this technique, which has been shown to be effective at avoiding detection. The separate corpus is transformed by paraphrasing the texts sentence by sentence. Both human and machine texts that are transformed get labeled as machine-generated.

We train PAWN with a \texttt{openai-community/\allowbreak gpt2}~\cite{radford2019language} and \texttt{meta-llama/\allowbreak Llama-3.2-1B-Instruct}~\cite{grattafiori2024Llama3} backbone, and compare it with the rest of the baselines on each of the testbeds selected by the authors. These baselines include FastText~\cite{joulin-etal-2017-bag}, GLTR~\cite{gehrmann2019GLTRStatisticala}, DetectGPT~\cite{mitchell2023DetectGPTZeroShot} and a fine-tuned LM for classification with a Longformer~\cite{beltagy2020LongformerLongDocument} backbone. \citeauthor{li2024MAGEMachinegenerateda} report having tested RoBERTa-\{base,large\}~\cite{liu2019RoBERTaRobustly}, BERT-\{base,large\}~\cite{devlin2019BERTPretraining} and GPT2-\{small,medium,large\}~\cite{radford2019language}, but finding Longformer to be the best performing one. We further include Binoculars~\cite{hans2024SpottingLLMs} as a zero-shot detector, and RADAR~\cite{hu2023radar} both in the zero-shot setting and with fine-tuning. It should be noted that the latter was trained in the WebText~\cite{radford2019language} corpus of scrapped documents, and out-of-distribution results should be taken with caution.

We adopted their evaluation metrics to be able to compare PAWN with their baselines. As the set is imbalanced, the macro-averaged recall is used. The AUROC is also reported, as sometimes the detectors are able to distinguish between machine and human text, but the exact threshold has a lot of variability depending on the underlying text distribution. All baselines in~\cite{li2024MAGEMachinegenerateda} are trained for five full epochs and we maintain their setting for a fair comparison.

\Cref{tab:mage-main-results-id,tab:mage-main-results-ood} show the results on MAGE's testbeds. We restrict our attention to the Longformer and RADAR-FT baselines, as they are the only competitive ones. In \Cref{tab:mcnemar} we show p-values in MAGE's main testbeds 4-8 from the McNemar test between PAWN models and our main baselines, except Longformer's testbeds 5 and 6 where checkpoints are not accessible. 

\begin{table}[!ht]
\centering
\begin{adjustbox}{max width=\textwidth}
\begin{tabular}{cccccc}
\toprule
\textbf{Setting}            & \textbf{Method}  & \textbf{HumanRec} & \textbf{MachineRec} & \textbf{AvgRec}  & \textbf{AUROC} \\
\midrule
\multicolumn{6}{c}{Testbed 2,3,4: In-distribution Detection}                                                                                        \\
\midrule
                                                   & FastText         & 88.96\%           & 77.08\%             & 83.02\%          & {\ul 0.89}     \\
                                                   & GLTR             & 75.61\%           & 79.56\%             & 77.58\%          & 0.84           \\
                                                   & Longformer       & 95.25\%           & 96.94\%             & \textbf{96.10\%} & \textbf{0.99}  \\
                                                   & DetectGPT*       & 48.67\%           & 75.95\%             & 62.31\%          & 0.60           \\
                                                   & RADAR*‡          & 52.25\%           & 66.25\%             & 59.25\%          & 0.60           \\
                                                   & RADAR-FT         & 90.23\%           & 97.13\%             & 93.68\%          & \textbf{0.99}  \\
                                                   & Binoculars*‡     & 10.56\%           & 7.86\%              & 9.21\%           & 0.67           \\
                                                   & PAWN (GPT2)      & 95.03\%           & 95.07\%             & 95.05\%          & \textbf{0.99}  \\
\multirow{-9}{*}{One source model \& all domains}  & PAWN (Llama3-1b) & 95.54\%           & 96.10\%             & {\ul 95.82\%}    & \textbf{0.99}  \\
\midrule
                                                   & FastText         & 89.43\%           & 73.91\%             & 81.67\%          & 0.89           \\
                                                   & GLTR             & 37.25\%           & 88.90\%             & 63.08\%          & 0.80           \\
                                                   & Longformer       & 89.78\%           & 97.24\%             & \textbf{93.51\%} & \textbf{0.99}  \\
                                                   & DetectGPT*       & 86.92\%           & 34.05\%             & 60.48\%          & 0.57           \\
                                                   & RADAR*‡          & 53.45\%           & 68.09\%             & 60.77\%          & 0.63           \\
                                                   & RADAR-FT         & 84.35\%           & 95.82\%             & 90.08\%          & 0.97           \\
                                                   & Binoculars*‡     & 10.22\%           & 8.35\%              & 9.28\%           & 0.65           \\
                                                   & PAWN (GPT2)      & 92.76\%           & 90.72\%             & 91.74\%          & {\ul 0.98}     \\
\multirow{-9}{*}{All source models \& one domain}  & PAWN (Llama3-1b) & 93.44\%           & 90.90\%             & {\ul 92.17\%}    & {\ul 0.98}     \\
\midrule
                                                   & FastText         & 86.34\%           & 71.26\%             & 78.80\%          & 0.83           \\
                                                   & GLTR             & 12.42\%           & 98.42\%             & 55.42\%          & 0.74           \\
                                                   & Longformer       & 82.80\%           & 98.27\%             & 90.53\%          & \textbf{0.99}  \\
                                                   & DetectGPT*       & 86.92\%           & 34.05\%             & 60.48\%          & 0.57           \\
                                                   & RADAR*‡          & 52.08\%           & 68.56\%             & 60.32\%          & 0.61           \\
                                                   & RADAR-FT         & 85.36\%           & 97.26\%             & 91.31\%          & {\ul 0.98}     \\
                                                   & Binoculars*‡     & 10.60\%           & 8.23\%              & 9.42\%           & 0.65           \\
                                                   & PAWN (GPT2)      & 93.75\%           & 91.97\%             & {\ul 92.86\%}    & {\ul 0.98}     \\
\multirow{-9}{*}{All source models \& all domains} & PAWN (Llama3-1b) & 90.68\%           & 95.84\%             & \textbf{93.26\%} & \textbf{0.99}  \\
\bottomrule
\end{tabular}
\end{adjustbox}
\caption{Results in MAGE's testbeds 2-4. Recalls for both classes and their macro-average are reported due to class imbalances, as well as the Area Under the ROC curve. The asterisk * denotes unsupervised models. The ‡ symbol denotes that the threshold has not been adjusted. All results except PAWN's, Binoculars' and RADAR's are taken from~\cite{li2024MAGEMachinegenerateda}.}
\label{tab:mage-main-results-id}
\end{table}

\begin{table}[!ht]
\centering
\begin{adjustbox}{max width=\textwidth}
\begin{tabular}{cccccc}
\toprule
Setting                                         & Method           & HumanRec & MachineRec & AvgRec  & AUROC \\
\midrule
\multicolumn{6}{c}{Testbed 5,6: Out-of-distribution Detection}                                                                                                                                           \\
\midrule
                                                 & FastText         & 83.12\%                         & 54.09\%                         & 68.61\%                         & 0.74                         \\
                                                 & GLTR             & 25.77\%                         & 89.21\%                         & 57.49\%                         & 0.65                         \\
                                                 & Longformer       & 83.31\%                         & 89.90\%                         & 86.61\%                         & 0.95                         \\
                                                 & DetectGPT*       & 48.67\%                         & 75.95\%                         & 62.31\%                         & 0.60                         \\
                                                 & RADAR*‡          & 52.25\%                         & 66.25\%                         & 59.25\%                         & 0.60                         \\
                                                 & RADAR-FT         & 80.71\%                         & 90.49\%                         & 85.60\%                         & 0.94                         \\
                                                 & Binoculars*‡     & 10.56\%                         & 7.86\%                          & 9.21\%                          & 0.67                         \\
                                                 & PAWN (GPT2)      & 93.97\%                         & 81.43\%                         & {\ul 87.70\%}                   & {\ul 0.96}                   \\
\multirow{-9}{*}{Unseen source model}            & PAWN (Llama3-1b) & 95.17\%                         & 85.68\%                         & \textbf{90.42\%}                & \textbf{0.97}                \\
\midrule
                                                 & FastText         & 54.29\%                         & 72.79\%                         & 63.54\%                         & 0.72                         \\
                                                 & GLTR             & 15.84\%                         & 97.12\%                         & 56.48\%                         & 0.72                         \\
                                                 & Longformer       & 38.05\%                         & 98.75\%                         & 68.40\%                         & 0.93                         \\
                                                 & DetectGPT*       & 86.92\%                         & 34.05\%                         & 60.48\%                         & 0.57                         \\
                                                 & RADAR*‡          & 53.45\%                         & 68.09\%                         & 60.77\%                         & 0.63                         \\
                                                 & RADAR-FT         & 38.81\% & 98.22\% & 68.52\%          & 0.85          \\
                                                 & Binoculars*‡     & 10.22\%                         & 8.35\%                          & 9.28\%                          & 0.65                         \\
                                                 & PAWN (GPT2)      & 65.17\%                         & 95.45\%                         & {\ul 80.31\%}                   & {\ul 0.94}                   \\
\multirow{-9}{*}{Unseen domain}                  & PAWN (Llama3-1b) & 67.24\%                         & 95.84\%                         & \textbf{81.54\%}                & \textbf{0.95}                \\
\midrule
\multicolumn{6}{c}{Testbed 7,8: Detection in the wilderness}                                                                                                                                             \\
\midrule
                                                 & Longformer       & 52.50\%                         & 99.14\%                         & 75.82\%                         & {\ul 0.94}                   \\
                                                 & Longformer †     & 88.78\%                         & 84.12\%                         & 86.54\%                         & {\ul 0.94}                   \\
                                                 & RADAR*‡          & 82.41\%                         & 73.62\%                         & 78.02\%                         & 0.87                         \\
                                                 & RADAR-FT         & 61.42\%                         & 98.12\%                         & 79.77\%                         & {\ul 0.94}                   \\
                                                 & Binoculars*‡     & 5.51\%                          & 7.12\%                          & 17.20\%                         & 0.88                         \\
                                                 & PAWN (GPT2)      & 84.12\%                         & 89.50\%                         & {\ul 86.81\%}                   & {\ul 0.94}                   \\
\multirow{-7}{*}{Unseen Domains \& Unseen Model} & PAWN (Llama3-1b) & 78.87\%                         & 96.75\%                         & \textbf{87.81\%}                & \textbf{0.97}                \\
\midrule
                                                 & Longformer       & 52.16\%                         & 81.73\%                         & {\ul 66.94\%}                   & {\ul 0.75}                   \\
                                                 & Longformer †     & 88.78\%                         & 37.05\%                         & 62.92\%                         & {\ul 0.75}                   \\
                                                 & RADAR*‡          & 67.67\%                         & 59.87\%                         & 63.77\%                         & 0.71                         \\
                                                 & RADAR-FT         & 53.33\%                         & 92.25\%                         & \textbf{72.79\%}                & \textbf{0.84}                \\
                                                 & Binoculars*‡     & 6.27\%                          & 16.00\%                         & 21.76\%                         & 0.55                         \\
                                                 & PAWN (GPT2)      & 87.00\%                         & 39.13\%                         & 63.06\%                         & {\ul 0.75}                   \\
\multirow{-7}{*}{Paraphrasing attacks}           & PAWN (Llama3-1b) & 77.53\%                         & 55.87\%                         & 66.70\%                         & 0.74                         \\
\bottomrule
\end{tabular}
\end{adjustbox}
\caption{Results in MAGE's testbeds 5-8. Recalls for both classes and their macro-average are reported due to class imbalances, as well as the Area Under the ROC curve. The asterisk * denotes unsupervised models. The dagger \dag  denotes that the classification boundary has been adjusted by using a portion of training data in each of the TB6 models and averaging the best boundaries. The ‡ symbol denotes that the threshold has not been adjusted. All results except PAWN's, Binoculars' and RADAR's are taken from~\cite{li2024MAGEMachinegenerateda}.}
\label{tab:mage-main-results-ood}
\end{table}

\begin{table}[!htpb]
    \centering
    \begin{adjustbox}{max width=\textwidth}
    \begin{tabular}{c|ccc|ccccc}
\toprule
           & \multicolumn{3}{c|}{Longformer}                             & \multicolumn{5}{c}{RADAR-FT}                                                      \\
\midrule
           & TB4                & TB7               & TB8               & TB4               & TB5       & TB6       & TB7               & TB8               \\
\midrule
PAWN-GPT2  & \textbf{\boldmath \(2.85 \cdot 10^{-100}\)} & \textbf{\boldmath \(3.37 \cdot 10^{-24}\)} & \textbf{\boldmath \(7.14 \cdot 10^{-20}\)} & \textbf{\boldmath \(5.37 \cdot 10^{-33}\)} & \textbf{\boldmath \(\approx 0\)} & \textbf{\boldmath \(\approx 0\)} & \textbf{\boldmath \(7.49 \cdot 10^{-10}\)} & \textbf{\boldmath \(1.60 \cdot 10^{-03}\)} \\
PAWN-LLaMA & \textbf{\boldmath \(3.77 \cdot 10^{-132}\)} & \textbf{\boldmath \(2.53 \cdot 10^{-37}\)} & \textbf{\boldmath \(2.18 \cdot 10^{-22}\)} & \textbf{\boldmath \(3.85 \cdot 10^{-51}\)} & \textbf{\boldmath \(\approx 0\)} & \textbf{\boldmath \(\approx 0\)} & \textbf{\boldmath \(1.32 \cdot 10^{-16}\)} & \textbf{\boldmath \(2.52 \cdot 10^{-03}\)} \\
\bottomrule
\end{tabular}
    \end{adjustbox}
    \caption{McNemar's p-values between PAWN models and the main baselines: Longformer (TB4,7,8) and RADAR-FT (TB4-8). Longformer TB5 and TB6 checkpoints are not available. In bold we show p-values under $0.05$, indicating a statistically significant difference in model performance.}
    \label{tab:mcnemar}
\end{table}

In the in-distribution settings of testbeds two and three, our model appears to perform on par, but slightly worse than Longformer, the strongest baseline. This is more pronounced when restricted to one domain than when restricted to one model. Thus, in the rare setting where we know the source model or the domain, we might get better performance with a fine-tuned LM. 

However, in the fourth test, without restrictions, both of our models outperform the Longformer and RADAR-FT baselines in terms of recall, by a good margin. The AUROC is slightly worse for PAWN-GPT2 than for Longformer, although the magnitude of the difference is hidden by the rounding. Still, PAWN models find a boundary that generalizes better in test data.

In out-of-distribution testbeds five and six, showing the ability to generalize to unseen domains and models, we find that both of our models outperform the baselines in the two metrics. Although the Longformer is very close in terms of AUROC, there is a big gap in average recall, especially in unseen domains. This again suggests that there is less variability in the decision boundary across distribution changes with PAWN models.

PAWN models also outperform in test seven for both metrics, with a good margin in the case of PAWN-LLaMA. Furthermore, the macro-averaged recall of RADAR-FT Longformer with the naive threshold is much weaker, and \citeauthor{li2024MAGEMachinegenerateda} had to use a special procedure to select a better one for the latter. Finally, the results in the eighth test suggest that our models are also not robust to paraphrasing attacks. The performance of PAWN is greatly diminished and cannot be used reliably in this setting. On the other hand, RADAR-FT performs much better than the other detectors. This is to be expected, as RADAR is trained adversarially against a paraphraser, specifically targeting this type of attacks. As this type of adversarial training can also be applied to PAWN, this might be an interesting way to improve performance against these attacks. We do not follow this line of research in this work.

We see a general trend across the results. In more specific in-distribution settings, our model performs slightly worse, but as we get to more open settings and even out-of-distribution settings, PAWN performs better, with the gap growing bigger. This is especially the case in terms of recall, which is dependent on the selected threshold, unlike AUROC. It appears that the PAWN detectors result in a more uniform decision boundary across domains and source models.

\subsection{Ablations}

To study the importance of each element in the Perplexity Attention Weighted Network, we run the MAGE experiments with each of the branches of our networks in isolation. We consider the following two variations.
\begin{enumerate}[label=(\roman*)]
    \item \textbf{Hidden state branch (HSFF).} We concatenate the consecutive hidden states and positional information, apply a single MLP element-wise, and average across tokens to produce the final prediction. We keep the number of hidden features in the MLPs and duplicate the depth of the network to match the maximum depth of computation.

    The HSFF model differs from PAWN by using semantic information directly for making predictions. PAWN, on the other hand, applies a softmax over the sequence to generate weights based on semantic and positional information, using them solely to modulate the influence of metric-based features for each token. This prevents semantic features from directly affecting the output, reducing the risk of semantic overfitting.
    
    \item \textbf{Metrics branch (MPN).} We remove the filtering based on hidden states and \emph{aggregate the metrics using a simple mean}. This differs from PAWN only in the aggregation scheme, as PAWN uses semantic and positional information to assign different weights to each token.
\end{enumerate}
For these ablations, we restrict ourselves to MAGE's testbeds 4-8, as they include the most difficult in-distribution setting and all out-of-distribution ones. \Cref{tab:mage-ablations-mixed} shows the results. For each backbone, the PAWN model outperforms the alternatives in all settings, with two exceptions. First, in the presence of paraphrasing attacks, where all models suffer greatly. Second, MPN has a slightly higher recall with the GPT2 backbone but a smaller AUROC.

\begin{table}[!ht]
\centering
\begin{adjustbox}{max width=\textwidth}
\begin{tabular}{cccccccc}
\toprule
\textbf{Setting}                                   & \textbf{Method}  & \textbf{HumanRec} & \textbf{MachineRec} & \textbf{AvgRec}  & \textbf{AUROC} & \textbf{AvgRec var} & \textbf{AUROC var} \\
\midrule
\multicolumn{8}{c}{Testbed 4: In-distribution Detection}                                                                                                                                                                       \\
\midrule
                                                   & PAWN (GPT2)      & 93.75\%           & 91.97\%             & {\ul 92.86\%}    & {\ul 0.981}    & \multicolumn{1}{l}{}                & \multicolumn{1}{l}{}               \\
                                                   & PAWN (Llama3-1b) & 90.68\%           & 95.84\%             & \textbf{93.26\%} & \textbf{0.986} & \multicolumn{1}{l}{}                & \multicolumn{1}{l}{}               \\
                                                   & HSFF (GPT2)      & 89.05\%           & 81.99\%             & 85.52\%          & 0.932          & \multicolumn{1}{l}{}                & \multicolumn{1}{l}{}               \\
                                                   & HSFF (Llama3-1b) & 94.57\%           & 83.18\%             & 88.87\%          & 0.960          & \multicolumn{1}{l}{}                & \multicolumn{1}{l}{}               \\
                                                   & MPN (GPT2)       & 83.02\%           & 82.64\%             & 82.83\%          & 0.926          & \multicolumn{1}{l}{}                & \multicolumn{1}{l}{}               \\
\multirow{-6}{*}{All source models \& all domains} & MPN (Llama3-1b)  & 85.49\%           & 81.84\%             & 83.67\%          & 0.927          & \multicolumn{1}{l}{}                & \multicolumn{1}{l}{}               \\
\midrule
\multicolumn{8}{c}{Testbed 5,6: Out-of-distribution Detection}                                                                                                                                                                 \\
\midrule
                                                   & PAWN (GPT2)      & 93.97\%           & 81.43\%             & {\ul 87.70\%}    & {\ul 0.957}    & {\color[HTML]{9C0006} -5.16\%}      & {\color[HTML]{9C0006} -0.024}      \\
                                                   & PAWN (Llama3-1b) & 95.17\%           & 85.68\%             & \textbf{90.42\%} & \textbf{0.970} & {\color[HTML]{9C0006} -2.83\%}      & {\color[HTML]{9C0006} -0.016}      \\
                                                   & HSFF (GPT2)      & 89.77\%           & 67.44\%             & 78.60\%          & 0.883          & {\color[HTML]{9C0006} -6.92\%}      & {\color[HTML]{9C0006} -0.049}      \\
                                                   & HSFF (Llama3-1b) & 94.60\%           & 74.61\%             & 84.60\%          & 0.937          & {\color[HTML]{9C0006} -4.27\%}      & {\color[HTML]{9C0006} -0.023}      \\
                                                   & MPN (GPT2)       & 86.00\%           & 72.89\%             & 79.45\%          & 0.911          & {\color[HTML]{9C0006} -3.38\%}      & {\color[HTML]{9C0006} -0.015}      \\
\multirow{-6}{*}{Unseen source model}              & MPN (Llama3-1b)  & 88.25\%           & 69.19\%             & 78.72\%          & 0.900          & {\color[HTML]{9C0006} -4.94\%}      & {\color[HTML]{9C0006} -0.027}      \\
\midrule
                                                   & PAWN (GPT2)      & 65.17\%           & 95.45\%             & 80.31\%          & {\ul 0.936}    & {\color[HTML]{9C0006} -12.55\%}     & {\color[HTML]{9C0006} -0.045}      \\
                                                   & PAWN (Llama3-1b) & 67.24\%           & 95.84\%             & \textbf{81.54\%} & \textbf{0.946} & {\color[HTML]{9C0006} -11.71\%}     & {\color[HTML]{9C0006} -0.040}      \\
                                                   & HSFF (GPT2)      & 53.85\%           & 88.12\%             & 70.98\%          & 0.829          & {\color[HTML]{9C0006} -14.54\%}     & {\color[HTML]{9C0006} -0.104}      \\
                                                   & HSFF (Llama3-1b) & 52.55\%           & 90.10\%             & 71.32\%          & 0.829          & {\color[HTML]{9C0006} -17.55\%}     & {\color[HTML]{9C0006} -0.131}      \\
                                                   & MPN (GPT2)       & 77.86\%           & 83.11\%             & 80.48\%          & 0.917          & {\color[HTML]{9C0006} -2.34\%}      & {\color[HTML]{9C0006} -0.008}      \\
\multirow{-6}{*}{Unseen domain}                    & MPN (Llama3-1b)  & 78.99\%           & 82.14\%             & {\ul 80.57\%}    & 0.906          & {\color[HTML]{9C0006} -3.10\%}      & {\color[HTML]{9C0006} -0.021}      \\
\midrule
\multicolumn{8}{c}{Testbed 7,8: Detection in the wilderness}                                                                                                                                                                   \\
\midrule
                                                   & PAWN (GPT2)      & 84.12\%           & 89.50\%             & {\ul 86.81\%}    & {\ul 0.942}    & {\color[HTML]{9C0006} -6.05\%}      & {\color[HTML]{9C0006} -0.039}      \\
                                                   & PAWN (Llama3-1b) & 78.87\%           & 96.75\%             & \textbf{87.81\%} & \textbf{0.973} & {\color[HTML]{9C0006} -5.45\%}      & {\color[HTML]{9C0006} -0.013}      \\
                                                   & HSFF (GPT2)      & 66.14\%           & 81.25\%             & 73.70\%          & 0.817          & {\color[HTML]{9C0006} -11.82\%}     & {\color[HTML]{9C0006} -0.116}      \\
                                                   & HSFF (Llama3-1b) & 71.78\%           & 88.88\%             & 80.33\%          & 0.871          & {\color[HTML]{9C0006} -8.54\%}      & {\color[HTML]{9C0006} -0.089}      \\
                                                   & MPN (GPT2)       & 86.22\%           & 75.63\%             & 80.92\%          & 0.884          & {\color[HTML]{9C0006} -1.91\%}      & {\color[HTML]{9C0006} -0.041}      \\
\multirow{-6}{*}{Unseen Domains \& Unseen Model}   & MPN (Llama3-1b)  & 91.60\%           & 77.25\%             & 84.43\%          & 0.938          & {\color[HTML]{92D050} +0.76\%}      & {\color[HTML]{92D050} +0.011}      \\
\midrule
                                                   & PAWN (GPT2)      & 87.00\%           & 39.13\%             & 63.06\%          & {\ul 0.755}    & {\color[HTML]{9C0006} -29.80\%}     & {\color[HTML]{9C0006} -0.226}      \\
                                                   & PAWN (Llama3-1b) & 77.53\%           & 55.87\%             & {\ul 66.70\%}    & 0.735          & {\color[HTML]{9C0006} -26.56\%}     & {\color[HTML]{9C0006} -0.251}      \\
                                                   & HSFF (GPT2)      & 65.75\%           & 71.00\%             & \textbf{68.37\%} & \textbf{0.756} & {\color[HTML]{9C0006} -17.14\%}     & {\color[HTML]{9C0006} -0.177}      \\
                                                   & HSFF (Llama3-1b) & 64.08\%           & 58.13\%             & 61.10\%          & 0.664          & {\color[HTML]{9C0006} -27.77\%}     & {\color[HTML]{9C0006} -0.296}      \\
                                                   & MPN (GPT2)       & 90.97\%           & 7.37\%              & 49.17\%          & 0.528          & {\color[HTML]{9C0006} -33.66\%}     & {\color[HTML]{9C0006} -0.398}      \\
\multirow{-6}{*}{Paraphrasing attacks}             & MPN (Llama3-1b)  & 94.05\%           & 15.75\%             & 54.90\%          & 0.688          & {\color[HTML]{9C0006} -28.77\%}     & {\color[HTML]{9C0006} -0.239}      \\
\bottomrule
\end{tabular}
\end{adjustbox}
\caption{Ablations in MAGE's testbeds 4-8. Recalls for both classes and their macro average are reported due to class imbalances, as well as the Area Under the ROC curve. We compare the performance of the Perplexity Attention Weighted Network (PAWN), the hidden state branch (HSFF) and the metrics branch (MPN). The last two columns show the variation of the macro-averaged recall and AUROC with respect to the in-distribution setting of testbed 4, calculated as the difference between the two values.}
\label{tab:mage-ablations-mixed}
\end{table}

An interesting pattern emerges in the performance of the models in different settings. We observe that the hidden state branch is clearly the better alternative in the in-distribution fourth test. In unseen models, they perform relatively on-par, with HSFF still slightly ahead in terms of recall. However, in unseen domains (tests six and seven), HSFF falls far behind MPN.


This phenomenon is more apparent in the last two columns of \Cref{tab:mage-ablations-mixed}, where we show the variation in performance between each out-of-distribution testbed and the fourth in-distribution testbed. MPN is the model that suffers the least, showing great generalization in terms of both AUROC and recall. HSFF is the one that suffers the most out-of-distribution on average. PAWN also suffers a bit more than MPN in new domains, with recall taking a bigger hit, suggesting greater threshold variance across domains.

We theorize that using the next-token distribution of LLMs generalizes greatly, as the backbones are pre-trained in a massive dataset and not overfitted to specific domains. Using semantic information as in HSFF seems to overfit the training distribution and generalize worse, as we might expect given that they learn semantic domain-specific information. PAWN is somewhere in the middle, since semantic information is only used to aggregate across tokens. This provides a greater ability to fit the model to in-distribution data while maintaining good generalization.


\subsection{Evaluating ensembles as a way to improve generalization}

One simple idea to improve the generalization in out-of-distribution settings is to ensemble different models. This is particularly fitting in PAWN, where we may train with a set of different backbones, each giving a different perspective from their unique tokenization and fine-tuning process. To evaluate this idea, we tested several simple ensembles, averaging the final logits, on MAGE's testbeds 4, 7 and 8.

In \Cref{tab:ensembles-mage} we show the results. We obtain minor performance gains in the in-distribution setting, although more pronounced in terms of recall than in terms of AUROC. In the out-of-distribution setting with unseen domains and models we find no AUROC improvement, only macro-averaged recall ones. Ensembles do not appear to provide any meaningful improvement in the presence of paraphrasing attacks, which remain very challenging.

Given these results, we recommend against using ensembled PAWN models, especially with different backbones, given the marginal differences in performance and the increased computational resource usage.

\begin{table}[!ht]
\centering
\begin{adjustbox}{max width=0.9\textwidth}
\begin{tabular}{cccccc}
\toprule
\textbf{Setting}                                  & \textbf{Backbones}          & \textbf{HumanRec} & \textbf{MachineRec} & \textbf{AvgRec}  & \textbf{AUROC} \\
\midrule
\multicolumn{6}{c}{Testbed 4: In-distribution Detection}                                                                                                      \\
\midrule
\multirow{7}{*}{All source models \& all domains} & GPT2                        & 93.75\%           & 91.97\%             & 92.86\%          & 0.981          \\
                                                  & LLaMA3-1b                   & 90.68\%           & 95.84\%             & 93.26\%          & 0.986          \\
                                                  & 1 GPT2, 1 LLaMA3-1b         & 94.59\%           & 93.92\%             & 94.26\%          & {\ul 0.988}    \\
                                                  & 2 GPT2, 2 LLaMA3-1b         & 95.86\%           & 93.55\%             & \textbf{94.71\%} & \textbf{0.989} \\
                                                  & 2 GPT2                      & 93.89\%           & 92.27\%             & 93.08\%          & 0.983          \\
                                                  & 2 LLaMA3-1b                 & 94.41\%           & 94.50\%             & {\ul    94.46\%} & {\ul 0.988}    \\
                                                  & 1 GPT2, 1 LLaMA3-1b, 1 QWEN & 90.78\%           & 96.13\%             & 93.45\%          & 0.987          \\
\midrule
\multicolumn{6}{c}{Testbed 7,8: Detection in the wilderness}                                                                                                  \\
\midrule
\multirow{7}{*}{Unseen Domains \& Unseen Model}   & GPT2                        & 84.12\%           & 89.50\%             & 86.81\%          & 0.942          \\
                                                  & LLaMA3-1b                   & 78.87\%           & 96.75\%             & 87.81\%          & \textbf{0.973} \\
                                                  & 1 GPT2, 1 LLaMA3-1b         & 84.78\%           & 94.00\%             & {\ul 89.39\%}    & 0.961          \\
                                                  & 2 GPT2, 2 LLaMA3-1b         & 85.70\%           & 93.75\%             & \textbf{89.72\%} & 0.963          \\
                                                  & 2 GPT2                      & 83.33\%           & 91.00\%             & 87.17\%          & 0.943          \\
                                                  & 2 LLaMA3-1b                 & 82.28\%           & 95.88\%             & 89.08\%          & {\ul 0.969}    \\
                                                  & 1 GPT2, 1 LLaMA3-1b, 1 QWEN & 82.28\%           & 94.50\%             & 88.39\%          & 0.954          \\
\midrule
\multirow{7}{*}{Paraphrasing attacks}             & GPT2                        & 87.00\%           & 39.13\%             & 63.06\%          & \textbf{0.755} \\
                                                  & LLaMA3-1b                   & 77.53\%           & 55.87\%             & \textbf{66.70\%} & 0.735          \\
                                                  & 1 GPT2, 1 LLaMA3-1b         & 86.49\%           & 42.00\%             & 64.25\%          & {\ul 0.751}    \\
                                                  & 2 GPT2, 2 LLaMA3-1b         & 86.62\%           & 39.50\%             & 63.06\%          & 0.739          \\
                                                  & 2 GPT2                      & 85.21\%           & 41.87\%             & 63.54\%          & 0.747          \\
                                                  & 2 LLaMA3-1b                 & 82.27\%           & 43.50\%             & 62.88\%          & 0.711          \\
                                                  & 1 GPT2, 1 LLaMA3-1b, 1 QWEN & 82.91\%           & 49.25\%             & {\ul 66.08\%}    & 0.742         \\
\bottomrule
\end{tabular}
\end{adjustbox}
\caption{Results from running several ensembles of our model, by simply averaging the final logits, on MAGE's testbeds four, seven and eight. Recalls for both classes and their macro average are reported due to class imbalances, as well as the Area Under the ROC curve.}
\label{tab:ensembles-mage}
\end{table}
\subsection{Generalization and multilingualism on M4}

Next, we evaluate the PAWN model in M4GT-Bench~\cite{wang2024M4GTBenchEvaluation} settings, and compare with their XLM-RoBERTa and RoBERTa baselines. These settings include leave-one-out experiments to test out-of-distribution performance, similar to the fifth and sixth testbeds in MAGE~\cite{li2024MAGEMachinegenerateda}. In particular, they divide the data by domain, generative model, or language, and evaluate detectors through cross-validation. We remark that their experiments across domains are performed as a multiclass problem, where the detector not only predicts whether the text is AI-generated, but also the source model. We maintain the binary classification setting and train our own RoBERTa~\cite{liu2019RoBERTaRobustly} base model as baseline.
We adopted their 10-epoch training limit, using only five epochs ourselves. We adapt to their metrics choice of accuracy, F1-score of the machine-generated class, and macro-averaged F1-score, depending on the case. Note that the accuracy metric is acceptable because the datasets are decently balanced.

\Cref{tab:results-m4gt} shows the results. In the \textbf{out-of-distribution domain tests} we find that the strongest model overall is PAWN-GPT2. Although in terms of AUROC it is a close race, with RoBERTa surpassing PAWN-LLaMA, both PAWN models outperform in terms of accuracy and macro-averaged F1. Thus, we find again that decision boundaries seem to generalize better in PAWN. In the \textbf{out-of-distribution language tests} the results are similar to those in the previous section, with PAWN-LLaMA generalizing nicely to other languages, although there is some performance degradation. PAWN-GPT2 does generalize to some of the languages, but still performs worse than XLM-RoBERTa. Finally, in \textbf{out-of-distribution model tests} the PAWN detectors outperform the baselines greatly.

\begin{table}[!ht]
    \centering
    \begin{adjustbox}{max width=\textwidth}
    \begin{tabular}{r|ccc|ccc|ccc|ccc}
\toprule
\toprule
           & \multicolumn{12}{c}{\textbf{OOD Domain}}                                                                                                                                                                 \\
\midrule
           & \multicolumn{3}{c}{PAWN (GPT2)}                      & \multicolumn{3}{c}{PAWN (LLaMA3-1b)}                 & \multicolumn{3}{c}{RoBERTa}                          &          &                  &       \\
\midrule
           & Accuracy         & F1-Macro         & AUROC          & Accuracy         & F1-Macro         & AUROC          & Accuracy         & F1-Macro         & AUROC          &          &                  &       \\
\midrule
Arxiv      & 95.97\%          & 95.96\%          & 0.988          & 88.25\%          & 88.17\%          & 0.995          & \textbf{99.23\%} & \textbf{99.23\%} & \textbf{1.000} &          &                  &       \\
Outfox     & 82.41\%          & 81.35\%          & 0.989          & 76.51\%          & 74.22\%          & 0.890          & \textbf{85.63\%} & \textbf{85.02\%} & \textbf{0.989} &          &                  &       \\
PeerRead   & \textbf{95.02\%} & \textbf{91.35\%} & 0.948          & 94.27\%          & 89.90\%          & \textbf{0.960} & 91.53\%          & 83.79\%          & 0.940          &          &                  &       \\
Reddit     & \textbf{92.11\%} & \textbf{92.04\%} & \textbf{0.982} & 91.97\%          & 91.95\%          & 0.975          & 82.11\%          & 81.92\%          & 0.958          &          &                  &       \\
Wikihow    & \textbf{94.10\%} & \textbf{94.08\%} & \textbf{0.982} & 90.49\%          & 90.49\%          & 0.965          & 86.17\%          & 86.17\%          & 0.942          &          &                  &       \\
Wikipedia  & 88.04\%          & 87.97\%          & 0.988          & \textbf{92.44\%} & \textbf{92.43\%} & 0.989          & 76.23\%          & 75.24\%          & \textbf{0.993} &          &                  &       \\
\midrule
Average    & \textbf{91.27\%} & \textbf{90.46\%} & \textbf{0.980} & 88.99\%          & 87.86\%          & 0.962          & 86.82\%          & 85.23\%          & 0.970          &          &                  &       \\
\midrule
\midrule
           & \multicolumn{12}{c}{\textbf{OOD Language}}                                                                                                                                                               \\
\midrule
           & \multicolumn{3}{c}{PAWN (GPT2)}                      & \multicolumn{3}{c}{PAWN (LLaMA3-1b)}                 & \multicolumn{3}{c}{XLM-RoBERTa}                      & \multicolumn{3}{c}{}                \\
\midrule
           & Accuracy         & F1-Macro         & AUROC          & Accuracy         & F1-Macro         & AUROC          & Accuracy         & F1-Macro         & AUROC          &          &                  &       \\
\midrule
Arabic     & 75.41\%          & 75.20\%          & 0.855          & 80.76\%          & 79.77\%          & 0.992          & \textbf{92.18\%} & \textbf{92.12\%} & -              &          &                  &       \\
Bulgarian  & 95.92\%          & 95.92\%          & 0.979          & \textbf{98.21\%} & \textbf{98.21\%} & 0.998          & 52.74\%          & 39.18\%          & -              &          &                  &       \\
Chinese    & 63.78\%          & 60.22\%          & 0.844          & 68.54\%          & 65.46\%          & 0.952          & \textbf{82.51\%} & \textbf{81.93\%} & -              &          &                  &       \\
English    & 59.11\%          & 58.09\%          & 0.695          & \textbf{76.71\%} & \textbf{76.59\%} & 0.839          & 66.51\%          & 64.55\%          & -              &          &                  &       \\
German     & 29.86\%          & 25.93\%          & 0.162          & \textbf{73.74\%} & \textbf{72.88\%} & 0.833          & 73.62\%          & 71.58\%          & -              &          &                  &       \\
Indonesian & 53.74\%          & 44.79\%          & 0.726          & \textbf{83.34\%} & \textbf{82.96\%} & 0.960          & 55.83\%          & 45.00\%          & -              &          &                  &       \\
Italian    & 62.37\%          & 57.53\%          & 0.783          & \textbf{96.43\%} & \textbf{96.43\%} & 0.994          & 83.51\%          & 83.04\%          & -              &          &                  &       \\
Russian    & \textbf{89.40\%} & \textbf{89.36\%} & 0.926          & 80.45\%          & 79.79\%          & 0.960          & 53.70\%          & 47.60\%          & -              &          &                  &       \\
Urdu       & 73.79\%          & 72.32\%          & 0.970          & 81.59\%          & 81.06\%          & 0.999          & \textbf{94.39\%} & \textbf{94.39\%} & -              &          &                  &       \\
\midrule
Average    & 67.04\%          & 64.37\%          & 0.771          & \textbf{82.20\%} & \textbf{81.46\%} & 0.947          & 72.78\%          & 68.82\%          & -              &          &                  &       \\
\midrule
\midrule
           & \multicolumn{12}{c}{\textbf{OOD Model}}                                                                                                                                                                  \\
\midrule
           & \multicolumn{3}{c}{PAWN (GPT2)}                      & \multicolumn{3}{c}{PAWN (LLaMA3-1b)}                 & \multicolumn{3}{c}{RoBERTa}                          & \multicolumn{3}{c}{XLM-RoBERTa}     \\
\midrule
           & Accuracy         & F1               & AUROC          & Accuracy         & F1               & AUROC          & Accuracy         & F1               & AUROC          & Accuracy & F1               & AUROC \\
\midrule
bloomz     & \textbf{88.36\%} & 56.86\%          & 0.847          & 87.32\%          & 48.57\%          & 0.858          & 60.30\%          & 60.22\%          & -              & 73.07\%  & \textbf{72.74\%} & -     \\
chatGPT    & 99.38\%          & 98.52\%          & 0.999          & \textbf{99.52\%} & \textbf{98.87\%} & 1.000          & 82.99\%          & 85.45\%          & -              & 85.62\%  & 87.57\%          & -     \\
cohere     & 99.01\%          & 97.10\%          & 0.999          & \textbf{99.09\%} & \textbf{97.32\%} & 0.998          & 78.24\%          & 81.91\%          & -              & 86.23\%  & 87.74\%          & -     \\
davinci    & 97.25\%          & 92.33\%          & 0.995          & \textbf{98.05\%} & \textbf{94.29\%} & 0.992          & 79.21\%          & 82.58\%          & -              & 84.32\%  & 85.23\%          & -     \\
dolly      & \textbf{96.72\%} & \textbf{90.23\%} & 0.990          & 96.25\%          & 88.18\%          & 0.994          & 77.78\%          & 81.44\%          & -              & 79.43\%  & 80.40\%          & -     \\
gpt4       & 99.46\%          & 98.55\%          & 1.000          & \textbf{99.67\%} & \textbf{99.12\%} & 1.000          & 79.37\%          & 82.90\%          & -              & 77.95\%  & 81.93\%          & -     \\
\midrule
Average    & \textbf{96.70\%} & \textbf{88.93\%} & 0.972          & 96.65\%          & 87.73\%          & 0.974          & 76.32\%          & 79.08\%          & -              & 81.10\%  & 82.60\%          & -     \\
\bottomrule
\bottomrule
\end{tabular}
    \end{adjustbox}
    \caption{Results in the M4 dataset~\cite{wang-etal-2024-m4,wang2024M4GTBenchEvaluation} for leave-one-out experiments across English domains (multilingual set), languages (multilingual set) and models (monolingual set). Results taken from~\cite{wang2024M4GTBenchEvaluation} do not report AUROC. Best results of comparable metrics are highlighted in bold.}
    \label{tab:results-m4gt}
\end{table}

\subsection{Generalization to new decoding strategies and robustness to attacks on RAID}

We conduct another set of experiments on RAID~\cite{dugan2024RAIDShareda}. First, we evaluate the ability of PAWN to generalize to a new domain and model selection, new decoding strategies, and the addition of repetition penalty. We do this by testing the PAWN models, the Longformer model from~\cite{li2024MAGEMachinegenerateda}, a RoBERTa-base~\cite{liu2019RoBERTaRobustly} model pre-trained on the main corpus of MAGE, and RADAR-PTM~\cite{hu2023radar} (RADAR pre-trained on the main corpus of MAGE). We also included baselines provided in RAID~\cite{dugan2024RAIDShareda}, but we note that each baseline is trained on a different dataset, and only zero-shot methods can be directly compared. These baselines include RoBERTa base and large models trained on the GPT2 output dataset~\cite{openai2024openaigpt2outputdataset}, another RoBERTa base model trained on the HC3 dataset~\cite{guo2023HowClose}, RADAR~\cite{hu2023radar}, GLTR~\cite{gehrmann2019GLTRStatisticala}, FastDetectGPT~\cite{bao2024fastdetectgpt}, Binoculars~\cite{hans2024SpottingLLMs}, and LLMDet~\cite{wu2023LLMDetThird}. They also included the closed-source commercial detectors GPTZero, Winston, Originality and ZeroGPT.

Second, we evaluate the model's ability to adapt to these new decoding settings and domain and model selection by fine-tuning it for one epoch on RAID's training set. Again, our main baselines are the Longformer from ~\cite{li2024MAGEMachinegenerateda}, the RoBERTa-base pre-trained on the main corpus of MAGE and RADAR-PTM (RADAR pre-trained on the main corpus of MAGE). We also add the base RADAR model. In this setting, no other baselines are provided in~\cite{dugan2024RAIDShareda}.

For uniformity with other baselines, we adopt the same metric that they used, that is, the recall at \(5\%\) FPR. Like them, we also select a different classification threshold per domain. It must also be noted that we do not use their test set, which is hidden, but our own split of their open data, which can be easily replicated from our code. We make sure to group texts that are generated from the same source title in the same sets of the split, avoiding any leakages.

\Cref{tab:raid-main-results} shows the main results on the RAID dataset. This table reports the recall at 5\% FPR by different categories of examples, depending on whether the model is open-source and instruction-tuned, the decoding strategy, and whether repetition penalty is used. To address each factor individually, we include \cref{tab:raid-main-results-variation}, where we group the values by each factor (e.g., all greedy decoding results vs. all sampling decoding results), take the median values and compute the difference. We remark that closed-source models are not included in the results without repetition penalty, as no corresponding counterparts are available.

Without fine-tuning, our models generalize slightly better than RoBERTa, and a lot better than Longformer and RADAR-PTM (pre-trained on MAGE), especially PAWN-LLaMA. Interestingly, in \Cref{tab:raid-main-results-variation} we find that our models are much more susceptible to the use of repetition penalty. Together with RADAR-PTM, PAWN models are also less susceptible to whether the model is instruction-tuned or not. One explanation for the first remark is that using repetition penalty tends to produce missing punctuation and stopwords at the end of the text, which results in very unexpected tokens in sharp distributions. This could easily mislead our models, which are based on these metrics.

After a single epoch of fine-tuning, we see that all models score close to perfect and are capable of adapting to the different factors without trouble.

One thing to note about these results is that the metric is not sensitive to threshold choice. This threshold is dynamically calculated to achieve the 5\% FPR, and, moreover, a different threshold is used for each domain. This hides the weakness we have been observing for fine-tuned LMs in other datasets.

\begin{table}[!ht]
\begin{adjustbox}{max width=\textwidth}
\begin{tabular}{r|cccccccccccc|c}
\toprule
                     & \multicolumn{8}{c|}{Open Source}                                                                                                                                                                                                                                                 & \multicolumn{4}{c|}{Closed Source}                                                                                                     & \multicolumn{1}{l}{Avg.}            \\
\toprule
                     & \multicolumn{4}{c|}{Chat}                                                                                                               & \multicolumn{4}{c|}{No Chat}                                                                                                            & \multicolumn{2}{c|}{Chat}                                          & \multicolumn{2}{c|}{No Chat}                                       & \multicolumn{1}{l}{}            \\
                     & \multicolumn{4}{c|}{(llama-c, mistral-c, mpt-c)}                                                                                        & \multicolumn{4}{c|}{(mistral, mpt, gpt2)}                                                                                               & \multicolumn{2}{c|}{(c-gpt, gpt4, cohere)}                         & \multicolumn{2}{c|}{(cohere, gpt3)}                                & \multicolumn{1}{l}{}            \\
\toprule
Decoding strategy    & \multicolumn{2}{c}{Greedy}                                        & \multicolumn{2}{c|}{Sampling}                                       & \multicolumn{2}{c}{Greedy}                                         & \multicolumn{2}{c|}{Sampling}                                      & \multicolumn{1}{c}{Greedy}      & \multicolumn{1}{c|}{Sampling}    & \multicolumn{1}{c}{Greedy}      & \multicolumn{1}{c|}{Sampling}    & \multicolumn{1}{l}{}            \\
Repetition Penalty   & \multicolumn{1}{c}{✗}           & \multicolumn{1}{c}{✓}           & \multicolumn{1}{c}{✗}            & \multicolumn{1}{c|}{✓}           & \multicolumn{1}{c}{✗}            & \multicolumn{1}{c}{✓}           & \multicolumn{1}{c}{✗}           & \multicolumn{1}{c|}{✓}           & \multicolumn{1}{c}{✗}           & \multicolumn{1}{c|}{✗}           & \multicolumn{1}{c}{✗}           & \multicolumn{1}{c|}{✗}           & \multicolumn{1}{l}{}            \\
\midrule
                     & \multicolumn{12}{c|}{No training on RAID}                                                                                                                                                                                                                                                                                                                                                                                &                                 \\
\midrule
R-B GPT2             & \cellcolor[HTML]{FFDA89}84.10\% & \cellcolor[HTML]{FFB264}52.30\% & \cellcolor[HTML]{FFD382}77.90\%  & \cellcolor[HTML]{FF9146}26.20\% & \cellcolor[HTML]{FFED9A}98.60\%  & \cellcolor[HTML]{FFA85A}44.10\% & \cellcolor[HTML]{FFBD6E}60.50\% & \cellcolor[HTML]{FF9D50}35.40\% & \cellcolor[HTML]{FFCA7A}70.90\% & \cellcolor[HTML]{FFA558}41.70\% & \cellcolor[HTML]{FFC273}65.10\% & \cellcolor[HTML]{FFB264}52.50\% & \cellcolor[HTML]{FFBB6C}59.11\% \\
R-L GPT2             & \cellcolor[HTML]{FFD584}79.70\% & \cellcolor[HTML]{FFA457}41.10\% & \cellcolor[HTML]{FFCA7A}71.40\%  & \cellcolor[HTML]{FF893E}19.50\% & \cellcolor[HTML]{FFED9A}98.50\%  & \cellcolor[HTML]{FFA659}43.00\% & \cellcolor[HTML]{FFC575}67.20\% & \cellcolor[HTML]{FFB465}53.40\% & \cellcolor[HTML]{FFBE6F}61.40\% & \cellcolor[HTML]{FF9C50}34.70\% & \cellcolor[HTML]{FFBD6E}61.10\% & \cellcolor[HTML]{FFAE60}48.60\% & \cellcolor[HTML]{FFB869}56.63\% \\
R-B CGPT             & \cellcolor[HTML]{FFD584}80.20\% & \cellcolor[HTML]{FFC071}63.30\% & \cellcolor[HTML]{FFD07F}75.50\%  & \cellcolor[HTML]{FFA255}39.30\% & \cellcolor[HTML]{FFB365}53.30\%  & \cellcolor[HTML]{FF9146}26.40\% & \cellcolor[HTML]{FF8338}14.90\% & \cellcolor[HTML]{FF7229}1.70\%  & \cellcolor[HTML]{FFBB6C}59.10\% & \cellcolor[HTML]{FFA053}38.10\% & \cellcolor[HTML]{FFAB5D}46.50\% & \cellcolor[HTML]{FFA155}39.00\% & \cellcolor[HTML]{FFA95B}44.78\% \\
RADAR                & \cellcolor[HTML]{FFE08E}88.80\% & \cellcolor[HTML]{FFD281}77.40\% & \cellcolor[HTML]{FFDC8B}85.60\%  & \cellcolor[HTML]{FFC474}66.40\% & \cellcolor[HTML]{FFE492}91.80\%  & \cellcolor[HTML]{FFC171}63.80\% & \cellcolor[HTML]{FFAD5F}48.30\% & \cellcolor[HTML]{FF984C}31.80\% & \cellcolor[HTML]{FFD786}81.60\% & \cellcolor[HTML]{FFCF7F}75.30\% & \cellcolor[HTML]{FFCB7B}72.20\% & \cellcolor[HTML]{FFC676}67.70\% & \cellcolor[HTML]{FFCA7A}70.89\% \\
GLTR                 & \cellcolor[HTML]{FFE290}89.80\% & \cellcolor[HTML]{FFC576}67.50\% & \cellcolor[HTML]{FFDA89}83.90\%  & \cellcolor[HTML]{FFA154}38.30\% & \cellcolor[HTML]{FFEE9B}99.60\%  & \cellcolor[HTML]{FFB869}56.90\% & \cellcolor[HTML]{FFA85B}44.50\% & \cellcolor[HTML]{FF7128}0.50\%  & \cellcolor[HTML]{FFD685}80.70\% & \cellcolor[HTML]{FFB566}54.30\% & \cellcolor[HTML]{FFD07F}75.60\% & \cellcolor[HTML]{FFC171}63.70\% & \cellcolor[HTML]{FFC070}62.94\% \\
F-DetectGPT          & \cellcolor[HTML]{FFED9A}98.60\% & \cellcolor[HTML]{FFCE7E}74.50\% & \cellcolor[HTML]{FFEA97}96.20\%  & \cellcolor[HTML]{FFA356}40.50\% & \cellcolor[HTML]{FFEC99}97.80\%  & \cellcolor[HTML]{FFB768}56.10\% & \cellcolor[HTML]{FFD584}79.70\% & \cellcolor[HTML]{FF7128}0.60\%  & \cellcolor[HTML]{FFE997}96.00\% & \cellcolor[HTML]{FFCE7D}74.10\% & \cellcolor[HTML]{FFE794}93.80\% & \cellcolor[HTML]{FFDD8C}86.30\% & \cellcolor[HTML]{FFCE7E}74.52\% \\
LLMDet               & \cellcolor[HTML]{FFB668}55.50\% & \cellcolor[HTML]{FF964A}30.20\% & \cellcolor[HTML]{FFAC5E}47.50\%  & \cellcolor[HTML]{FF853A}16.50\% & \cellcolor[HTML]{FFCF7E}74.80\%  & \cellcolor[HTML]{FF9247}27.00\% & \cellcolor[HTML]{FFA154}38.40\% & \cellcolor[HTML]{FF752B}3.70\%  & \cellcolor[HTML]{FF9D51}35.80\% & \cellcolor[HTML]{FF883D}18.50\% & \cellcolor[HTML]{FFA356}40.00\% & \cellcolor[HTML]{FF9A4D}32.90\% & \cellcolor[HTML]{FF9C50}35.07\% \\
Binoculars           & \cellcolor[HTML]{FFEE9B}99.90\% & \cellcolor[HTML]{FFDE8C}86.60\% & \cellcolor[HTML]{FFEE9B}99.70\%  & \cellcolor[HTML]{FFBD6E}60.60\% & \cellcolor[HTML]{FFEE9B}99.90\%  & \cellcolor[HTML]{FFBF70}62.30\% & \cellcolor[HTML]{FFCC7B}72.40\% & \cellcolor[HTML]{FF7128}0.60\%  & \cellcolor[HTML]{FFED9B}99.20\% & \cellcolor[HTML]{FFE592}92.10\% & \cellcolor[HTML]{FFED9A}99.00\% & \cellcolor[HTML]{FFE896}95.00\% & \cellcolor[HTML]{FFD685}80.61\% \\
GPTZero              & \cellcolor[HTML]{FFED9A}98.80\% & \cellcolor[HTML]{FFE794}93.70\% & \cellcolor[HTML]{FFEC9A}98.40\%  & \cellcolor[HTML]{FFD887}82.50\% & \cellcolor[HTML]{FFCF7E}74.70\%  & \cellcolor[HTML]{FF9C4F}34.60\% & \cellcolor[HTML]{FF7C32}9.40\%  & \cellcolor[HTML]{FF762D}4.80\%  & \cellcolor[HTML]{FFE593}92.30\% & \cellcolor[HTML]{FFE08E}88.50\% & \cellcolor[HTML]{FFBD6E}60.60\% & \cellcolor[HTML]{FFB465}53.40\% & \cellcolor[HTML]{FFC374}65.98\% \\
Originality          & \cellcolor[HTML]{FFED9A}98.60\% & \cellcolor[HTML]{FFDD8C}86.30\% & \cellcolor[HTML]{FFEC99}97.70\%  & \cellcolor[HTML]{FFCC7C}72.50\% & \cellcolor[HTML]{FFEE9B}99.90\%  & \cellcolor[HTML]{FFC172}64.10\% & \cellcolor[HTML]{FFE18F}89.00\% & \cellcolor[HTML]{FFB163}51.20\% & \cellcolor[HTML]{FFEA98}96.80\% & \cellcolor[HTML]{FFE18F}89.00\% & \cellcolor[HTML]{FFE492}91.70\% & \cellcolor[HTML]{FFDC8B}85.40\% & \cellcolor[HTML]{FFDC8A}85.18\% \\
Winston              & \cellcolor[HTML]{FFEB98}97.20\% & \cellcolor[HTML]{FFE290}90.10\% & \cellcolor[HTML]{FFEA98}96.60\%  & \cellcolor[HTML]{FFD382}78.30\% & \cellcolor[HTML]{FFC677}68.20\%  & \cellcolor[HTML]{FFAE60}49.00\% & \cellcolor[HTML]{FF9549}29.50\% & \cellcolor[HTML]{FF7E34}11.30\% & \cellcolor[HTML]{FFEA97}96.10\% & \cellcolor[HTML]{FFE794}93.70\% & \cellcolor[HTML]{FFCD7C}73.20\% & \cellcolor[HTML]{FFC676}68.10\% & \cellcolor[HTML]{FFCA7A}70.94\% \\
ZeroGPT*             & \cellcolor[HTML]{FFE996}95.40\% & \cellcolor[HTML]{FFD685}80.70\% & \cellcolor[HTML]{FFE290}90.50\%  & \cellcolor[HTML]{FFB667}54.90\% & \cellcolor[HTML]{FFDC8A}85.10\%  & \cellcolor[HTML]{FFB86A}57.20\% & \cellcolor[HTML]{FF843A}16.00\% & \cellcolor[HTML]{FF7128}0.30\%  & \cellcolor[HTML]{FFE592}92.10\% & \cellcolor[HTML]{FFC374}65.80\% & \cellcolor[HTML]{FFDA88}83.40\% & \cellcolor[HTML]{FFCC7C}72.70\% & \cellcolor[HTML]{FFC474}66.18\% \\
\midrule
PAWN (GPT2)          & \cellcolor[HTML]{FFEA97}96.56\% & \cellcolor[HTML]{FFE08E}88.57\% & \cellcolor[HTML]{FFE794}93.70\%  & \cellcolor[HTML]{FFD483}79.30\% & \cellcolor[HTML]{FFE390}90.53\%  & \cellcolor[HTML]{FFDA89}84.03\% & \cellcolor[HTML]{FFB96A}57.52\% & \cellcolor[HTML]{FFD786}81.56\% & \cellcolor[HTML]{FFE290}90.16\% & \cellcolor[HTML]{FFCB7B}72.02\% & \cellcolor[HTML]{FFDB89}84.34\% & \cellcolor[HTML]{FFCF7F}75.37\% & \cellcolor[HTML]{FFD987}82.81\% \\
PAWN (Llama3-1b)     & \cellcolor[HTML]{FFEC9A}98.33\% & \cellcolor[HTML]{FFE492}92.00\% & \cellcolor[HTML]{FFEA97}96.26\%  & \cellcolor[HTML]{FFD786}81.61\% & \cellcolor[HTML]{FFED9A}98.90\%  & \cellcolor[HTML]{FFDE8C}86.85\% & \cellcolor[HTML]{FFB264}52.07\% & \cellcolor[HTML]{FFD887}82.44\% & \cellcolor[HTML]{FFE693}93.10\% & \cellcolor[HTML]{FFCD7C}73.24\% & \cellcolor[HTML]{FFD988}83.15\% & \cellcolor[HTML]{FFCF7E}75.04\% & \cellcolor[HTML]{FFDB89}84.42\% \\
Longformer           & \cellcolor[HTML]{FFDF8D}87.89\% & \cellcolor[HTML]{FFDE8C}86.60\% & \cellcolor[HTML]{FFDD8C}86.55\%  & \cellcolor[HTML]{FFDA88}83.43\% & \cellcolor[HTML]{FFDD8B}86.15\%  & \cellcolor[HTML]{FFDA89}83.78\% & \cellcolor[HTML]{FFBB6D}59.62\% & \cellcolor[HTML]{FFDA88}83.63\% & \cellcolor[HTML]{FFD786}81.66\% & \cellcolor[HTML]{FFC677}68.24\% & \cellcolor[HTML]{FFC172}64.13\% & \cellcolor[HTML]{FFB566}54.11\% & \cellcolor[HTML]{FFD281}77.15\% \\
RoBERTa              & \cellcolor[HTML]{FFE290}89.89\% & \cellcolor[HTML]{FFEC99}98.10\% & \cellcolor[HTML]{FFE391}91.14\%  & \cellcolor[HTML]{FFE290}90.26\% & \cellcolor[HTML]{FFDC8A}85.32\%  & \cellcolor[HTML]{FFE996}95.55\% & \cellcolor[HTML]{FFAE60}49.14\% & \cellcolor[HTML]{FFE18F}89.24\% & \cellcolor[HTML]{FFE18F}89.04\% & \cellcolor[HTML]{FFCB7B}71.97\% & \cellcolor[HTML]{FFC777}68.53\% & \cellcolor[HTML]{FFC474}66.02\% & \cellcolor[HTML]{FFD887}82.02\% \\
RADAR-PTM            & \cellcolor[HTML]{FFE08E}88.14\% & \cellcolor[HTML]{FFDE8C}87.05\% & \cellcolor[HTML]{FFDD8B}85.80\%  & \cellcolor[HTML]{FFD887}82.54\% & \cellcolor[HTML]{FFE795}94.10\%  & \cellcolor[HTML]{FFE08E}88.66\% & \cellcolor[HTML]{FFAE60}49.15\% & \cellcolor[HTML]{FFE492}92.05\% & \cellcolor[HTML]{FFDA88}83.51\% & \cellcolor[HTML]{FFC676}67.96\% & \cellcolor[HTML]{FFC474}66.29\% & \cellcolor[HTML]{FFB869}56.80\% & \cellcolor[HTML]{FFD382}78.50\% \\
\midrule
\multicolumn{1}{l|}{} & \multicolumn{12}{c|}{1 epoch fine-tune on RAID}                                                                                                                                                                                                                                                                                                                                                                          &                                 \\
\midrule
PAWN (GPT2)          & \cellcolor[HTML]{FFEE9B}99.90\% & \cellcolor[HTML]{FFEE9B}99.65\% & \cellcolor[HTML]{FFEE9B}99.85\%  & \cellcolor[HTML]{FFEE9B}99.60\% & \cellcolor[HTML]{FFEE9B}99.98\%  & \cellcolor[HTML]{FFEE9B}99.83\% & \cellcolor[HTML]{FFEC99}98.23\% & \cellcolor[HTML]{FFEE9B}99.63\% & \cellcolor[HTML]{FFED9A}99.08\% & \cellcolor[HTML]{FFEC99}97.73\% & \cellcolor[HTML]{FFED9A}98.69\% & \cellcolor[HTML]{FFEA98}96.60\% & \cellcolor[HTML]{FFED9A}99.06\% \\
PAWN (Llama3-1b)     & \cellcolor[HTML]{FFEE9B}99.98\% & \cellcolor[HTML]{FFEE9B}99.98\% & \cellcolor[HTML]{FFEF9C}100.00\% & \cellcolor[HTML]{FFEE9B}99.93\% & \cellcolor[HTML]{FFEF9C}100.00\% & \cellcolor[HTML]{FFEE9B}99.95\% & \cellcolor[HTML]{FFEE9B}99.43\% & \cellcolor[HTML]{FFEE9B}99.93\% & \cellcolor[HTML]{FFEE9B}99.65\% & \cellcolor[HTML]{FFED9A}98.93\% & \cellcolor[HTML]{FFEE9B}99.63\% & \cellcolor[HTML]{FFED9A}98.84\% & \cellcolor[HTML]{FFEE9B}99.69\% \\
Longformer           & \cellcolor[HTML]{FFEE9B}99.78\% & \cellcolor[HTML]{FFEE9B}99.75\% & \cellcolor[HTML]{FFEE9B}99.65\%  & \cellcolor[HTML]{FFEE9B}99.55\% & \cellcolor[HTML]{FFEE9B}99.90\%  & \cellcolor[HTML]{FFED9A}99.10\% & \cellcolor[HTML]{FFEC99}97.88\% & \cellcolor[HTML]{FFEE9B}99.30\% & \cellcolor[HTML]{FFED9B}99.15\% & \cellcolor[HTML]{FFED9A}98.53\% & \cellcolor[HTML]{FFEC99}98.21\% & \cellcolor[HTML]{FFEA97}96.45\% & \cellcolor[HTML]{FFED9A}98.94\% \\
RoBERTa              & \cellcolor[HTML]{FFEE9B}99.98\% & \cellcolor[HTML]{FFEE9B}99.95\% & \cellcolor[HTML]{FFEE9B}99.93\%  & \cellcolor[HTML]{FFEE9B}99.83\% & \cellcolor[HTML]{FFEE9B}99.93\%  & \cellcolor[HTML]{FFEE9B}99.80\% & \cellcolor[HTML]{FFED9A}98.88\% & \cellcolor[HTML]{FFEE9B}99.35\% & \cellcolor[HTML]{FFEE9B}99.43\% & \cellcolor[HTML]{FFED9B}99.20\% & \cellcolor[HTML]{FFED9A}99.14\% & \cellcolor[HTML]{FFED9A}98.65\% & \cellcolor[HTML]{FFEE9B}99.50\% \\
RADAR-PTM            & \cellcolor[HTML]{FFEE9B}99.70\% & \cellcolor[HTML]{FFEE9B}99.58\% & \cellcolor[HTML]{FFEE9B}99.60\%  & \cellcolor[HTML]{FFED9A}99.10\% & \cellcolor[HTML]{FFEE9B}99.68\%  & \cellcolor[HTML]{FFEE9B}99.40\% & \cellcolor[HTML]{FFEB99}97.61\% & \cellcolor[HTML]{FFEB99}97.46\% & \cellcolor[HTML]{FFEE9B}99.40\% & \cellcolor[HTML]{FFED9B}99.18\% & \cellcolor[HTML]{FFEC9A}98.36\% & \cellcolor[HTML]{FFEB98}97.20\% & \cellcolor[HTML]{FFED9A}98.86\% \\
\midrule
RADAR                & \cellcolor[HTML]{FFE18F}89.61\% & \cellcolor[HTML]{FFE290}90.06\% & \cellcolor[HTML]{FFE18F}89.54\%  & \cellcolor[HTML]{FFE290}90.26\% & \cellcolor[HTML]{FFE08E}88.54\%  & \cellcolor[HTML]{FFDE8C}87.05\% & \cellcolor[HTML]{FFDC8B}85.73\% & \cellcolor[HTML]{FFDE8C}86.95\% & \cellcolor[HTML]{FFE391}91.23\% & \cellcolor[HTML]{FFE290}90.48\% & \cellcolor[HTML]{FFE391}90.81\% & \cellcolor[HTML]{FFE592}92.23\% & \cellcolor[HTML]{FFE18F}89.37\% \\
\bottomrule
\end{tabular}
\end{adjustbox}
\caption{Recall at 5\% FPR on the RAID dataset. Results are divided by several factors: whether the model is open source, whether it is a model fine-tuned for chat interactions, the decoding strategy and the use of repetition penalty during generation. Models denoted by an asterisk * did not achieve the target FPR. PAWN models, Longformer, RADAR-PTM (pre-trained on MAGE), RADAR fine-tuned on RAID, and RoBERTa were evaluated on our data split, as the original test split is hidden. The results from other detectors are taken from~\cite{dugan2024RAIDShareda}. These detectors were pre-trained on different training sets, and they were tested on the hidden set.}
\label{tab:raid-main-results}
\end{table}

\begin{table}[!ht]
\centering
\begin{adjustbox}{max width=0.8\textwidth}
\begin{tabular}{r|ccc}
\toprule
                     & Sampling vs. Greedy             & RP vs. No RP                    & No Chat vs. Chat                \\
\midrule
                     & \multicolumn{3}{c}{No training on RAID}                                                             \\
\midrule
R-B GPT2             & {\color[HTML]{9C0006} -15.60\%} & {\color[HTML]{9C0006} -41.25\%} & {\color[HTML]{9C0006} -5.10\%}  \\
R-L GPT2             & {\color[HTML]{9C0006} -15.45\%} & {\color[HTML]{9C0006} -33.50\%} & {\color[HTML]{4EA72E} +6.00\%}  \\
R-B CGPT             & {\color[HTML]{9C0006} -17.65\%} & {\color[HTML]{9C0006} -31.55\%} & {\color[HTML]{9C0006} -28.50\%} \\
RADAR                & {\color[HTML]{9C0006} -8.00\%}  & {\color[HTML]{9C0006} -22.10\%} & {\color[HTML]{9C0006} -13.75\%} \\
GLTR                 & {\color[HTML]{9C0006} -17.85\%} & {\color[HTML]{9C0006} -39.25\%} & {\color[HTML]{9C0006} -13.80\%} \\
F-DetectGPT          & {\color[HTML]{9C0006} -17.80\%} & {\color[HTML]{9C0006} -48.70\%} & {\color[HTML]{9C0006} -2.25\%}  \\
LLMDet               & {\color[HTML]{9C0006} -6.35\%}  & {\color[HTML]{9C0006} -29.75\%} & {\color[HTML]{4EA72E} +2.65\%}  \\
Binoculars           & {\color[HTML]{9C0006} -9.75\%}  & {\color[HTML]{9C0006} -38.35\%} & {\color[HTML]{9C0006} -11.95\%} \\
GPTZero              & {\color[HTML]{9C0006} -12.55\%} & {\color[HTML]{9C0006} -28.00\%} & {\color[HTML]{9C0006} -49.00\%} \\
Originality          & {\color[HTML]{9C0006} -6.60\%}  & {\color[HTML]{9C0006} -29.85\%} & {\color[HTML]{9C0006} -5.70\%}  \\
Winston              & {\color[HTML]{9C0006} -2.55\%}  & {\color[HTML]{9C0006} -18.75\%} & {\color[HTML]{9C0006} -36.35\%} \\
ZeroGPT*             & {\color[HTML]{9C0006} -15.00\%} & {\color[HTML]{9C0006} -31.75\%} & {\color[HTML]{9C0006} -20.65\%} \\
\midrule
PAWN (GPT2)          & {\color[HTML]{9C0006} -9.66\%}  & {\color[HTML]{9C0006} -9.32\%}  & {\color[HTML]{9C0006} -6.56\%}  \\
PAWN (Llama3-1b)     & {\color[HTML]{9C0006} -11.61\%} & {\color[HTML]{9C0006} -12.66\%} & {\color[HTML]{9C0006} -9.76\%}  \\
Longformer           & {\color[HTML]{9C0006} -8.96\%}  & {\color[HTML]{9C0006} -2.64\%}  & {\color[HTML]{9C0006} -11.11\%} \\
RoBERTa              & {\color[HTML]{9C0006} -7.91\%}  & {\color[HTML]{4EA72E} +5.31\%}  & {\color[HTML]{9C0006} -13.15\%} \\
RADAR-PTM            & {\color[HTML]{9C0006} -10.71\%} & {\color[HTML]{4EA72E} +0.88\%}  & {\color[HTML]{9C0006} -7.17\%}  \\
\midrule
\multicolumn{1}{l}{} & \multicolumn{3}{|c}{{1 epoch fine-tune on RAID}}                                \\
\midrule
PAWN (GPT2)          & {\color[HTML]{9C0006} -0.87\%}  & {\color[HTML]{4EA72E} +0.19\%}  & {\color[HTML]{9C0006} -0.48\%}  \\
PAWN (Llama3-1b)     & {\color[HTML]{9C0006} -0.34\%}  & {\color[HTML]{4EA72E} +0.09\%}  & {\color[HTML]{9C0006} -0.11\%}  \\
Longformer           & {\color[HTML]{9C0006} -0.75\%}  & {\color[HTML]{4EA72E} +0.12\%}  & {\color[HTML]{9C0006} -0.93\%}  \\
RoBERTa              & {\color[HTML]{9C0006} -0.30\%}  & {\color[HTML]{4EA72E} +0.06\%}  & {\color[HTML]{9C0006} -0.43\%}  \\
RADAR-PTM            & {\color[HTML]{9C0006} -0.59\%}  & {\color[HTML]{9C0006} -0.26\%}  & {\color[HTML]{9C0006} -1.14\%}  \\
\midrule
RADAR                & {\color[HTML]{9C0006} -0.37\%}  & {\color[HTML]{4EA72E} +0.22\%}  & {\color[HTML]{9C0006} -1.65\%}  \\
\bottomrule
\end{tabular}

\end{adjustbox}
\caption{Variation of recall at 5\% FPR on the RAID dataset, by different factors. We report the differences between the median values of each group of results. Closed-source models are not taken into account in the no repetition penalty category, as no corresponding counterparts are available. PAWN models, Longformer, RADAR-PTM (pre-trained on MAGE), RADAR fine-tuned on RAID, and RoBERTa were evaluated on our data split, as the original test split is hidden. The results from other detectors are taken from~\cite{dugan2024RAIDShareda}. These detectors were pre-trained on different training sets, and they were tested on the hidden set.}
\label{tab:raid-main-results-variation}
\end{table}

Finally, we evaluate the performance drop when different adversarial attacks are applied. This is tested on models that were fine-tuned on RAID's training set, excluding attacked data. The results appear on \Cref{tab:raid-attack-results}.

Let us consider first results in the second (pre-trained on MAGE and fine-tuned on RAID) and third (fine-tuned on RAID) boxes. We observe that on average, the PAWN models are the most robust (among models trained under the same conditions). PAWN-LLaMA seems particularly robust, with paraphrasing attacks being the only significantly damaging ones. Longformer, RoBERTa, RADAR, RADAR-PTM (pre-trained on MAGE) and PAWN-GPT2 are also susceptible to the addition of homoglyphs, the latter to a lesser extent. Zero-width spaces seem to affect only our fine-tuned LM selection, although RADAR-PTM to a lesser degree. The addition of extra whitespace only affects PAWN-GPT2 significantly.

Interestingly, RADAR and RADAR-PTM are the worst performing models in the presence of adversarial attacks, despite having been trained adversarially against these attacks. While it is possible that these capabilities were lost during fine-tuning, these results still contrast with those on MAGE's testbed 8.

The methods in the first box are not fine-tuned on RAID, but trained on different datasets. They are generally inferior in absolute terms. The notable exception is the great performance of Originality on paraphrasing attacks, at least in this evaluation set. Still, comparisons must be taken with a grain of salt due to the difference in test set selection and the fact that right-side models are fine-tuned on RAID.

\begin{table}[!ht]
\begin{adjustbox}{max width=\textwidth}
\begin{tabular}{c|cccccc|ccccc|c}
\toprule
                   & R-L GPT2                              & RADAR                           & GLTR                            & Binoculars                           & GPTZero                              & Originality                           & PAWN (GPT2)                     & PAWN (LLaMA)                    & Longformer                      & RoBERTa                         & RADAR-PTM                       & RADAR                           \\
\midrule
None                                   & {\ul 56.70\%}                         & 65.61\%                         & 59.69\%                         & 78.98\%                              & {\ul 66.50\%}                        & {\ul 85.00\%}                         & 99.15\%                         & \textbf{99.71\%}                & 99.03\%                         & 99.54\%                         & 98.92\%                         & 89.25\%                         \\
\midrule
                                       & {\ul 40.10\%}                         & 61.14\%                         & 43.06\%                         & 68.69\%                              & {\ul 66.20\%}                        & {\ul 84.90\%}                         & 93.76\%                         & \textbf{98.95\%}                & 97.44\%                         & 97.63\%                         & 96.90\%                         & 93.28\%                         \\
\multirow{-2}{*}{Whitespace}           & {\color[HTML]{9C0006} {\ul -16.60\%}} & {\color[HTML]{9C0006} -4.48\%}  & {\color[HTML]{9C0006} -16.64\%} & {\color[HTML]{9C0006} -10.30\%}      & {\color[HTML]{9C0006} {\ul -0.30\%}} & {\color[HTML]{9C0006} {\ul -0.10\%}}  & {\color[HTML]{9C0006} -5.39\%}  & {\color[HTML]{9C0006} -0.76\%}  & {\color[HTML]{9C0006} -1.59\%}  & {\color[HTML]{9C0006} -1.91\%}  & {\color[HTML]{9C0006} -2.02\%}  & {\color[HTML]{4EA72E} +4.04\%}  \\
                                       & -                                     & 65.14\%                         & 45.35\%                         & 72.85\%                              & -                                    & -                                     & 97.80\%                         & 98.86\%                         & 98.35\%                         & \textbf{99.25\%}                & 96.88\%                         & 90.37\%                         \\
\multirow{-2}{*}{Upper-Lower}          & {\color[HTML]{4EA72E} -}              & {\color[HTML]{9C0006} -0.47\%}  & {\color[HTML]{9C0006} -14.35\%} & {\color[HTML]{9C0006} -6.14\%}       & {\color[HTML]{4EA72E} -}             & {\color[HTML]{4EA72E} -}              & {\color[HTML]{9C0006} -1.34\%}  & {\color[HTML]{9C0006} -0.85\%}  & {\color[HTML]{9C0006} -0.69\%}  & {\color[HTML]{9C0006} -0.29\%}  & {\color[HTML]{9C0006} -2.04\%}  & {\color[HTML]{4EA72E} +1.12\%}  \\
                                       & {\ul 79.40\%}                         & 62.74\%                         & 28.74\%                         & 42.06\%                              & {\ul 61.00\%}                        & {\ul 96.50\%}                         & 97.77\%                         & 98.03\%                         & 98.18\%                         & \textbf{99.25\%}                & 97.57\%                         & 87.88\%                         \\
\multirow{-2}{*}{Synonym}              & {\color[HTML]{4EA72E} {\ul +22.70\%}} & {\color[HTML]{9C0006} -2.87\%}  & {\color[HTML]{9C0006} -30.95\%} & {\color[HTML]{9C0006} -36.92\%}      & {\color[HTML]{9C0006} {\ul -5.50\%}} & {\color[HTML]{4EA72E} {\ul +11.50\%}} & {\color[HTML]{9C0006} -1.37\%}  & {\color[HTML]{9C0006} -1.68\%}  & {\color[HTML]{9C0006} -0.85\%}  & {\color[HTML]{9C0006} -0.29\%}  & {\color[HTML]{9C0006} -1.35\%}  & {\color[HTML]{9C0006} -1.37\%}  \\
                                       & {\ul 39.50\%}                         & 64.31\%                         & 56.99\%                         & 77.25\%                              & {\ul 65.10\%}                        & {\ul 78.60\%}                         & 98.56\%                         & \textbf{99.32\%}                & 98.60\%                         & 99.27\%                         & 98.57\%                         & 90.73\%                         \\
\multirow{-2}{*}{Misspelling}          & {\color[HTML]{9C0006} {\ul -17.20\%}} & {\color[HTML]{9C0006} -1.30\%}  & {\color[HTML]{9C0006} -2.70\%}  & {\color[HTML]{9C0006} -1.74\%}       & {\color[HTML]{9C0006} {\ul -1.40\%}} & {\color[HTML]{9C0006} {\ul -6.40\%}}  & {\color[HTML]{9C0006} -0.58\%}  & {\color[HTML]{9C0006} -0.39\%}  & {\color[HTML]{9C0006} -0.43\%}  & {\color[HTML]{9C0006} -0.27\%}  & {\color[HTML]{9C0006} -0.35\%}  & {\color[HTML]{4EA72E} +1.49\%}  \\
                                       & {\ul 72.90\%}                         & 62.36\%                         & 42.95\%                         & 80.30\%                              & {\ul 64.00\%}                        & {\ul 96.70\%}                         & 85.30\%                         & 85.80\%                         & \textbf{88.48\%}                & 87.45\%                         & 84.61\%                         & 84.81\%                         \\
\multirow{-2}{*}{Paraphrase}           & {\color[HTML]{4EA72E} {\ul +16.20\%}} & {\color[HTML]{9C0006} -3.26\%}  & {\color[HTML]{9C0006} -16.74\%} & {\color[HTML]{4EA72E} {\ul +1.32\%}} & {\color[HTML]{9C0006} {\ul -2.50\%}} & {\color[HTML]{4EA72E} {\ul +11.70\%}} & {\color[HTML]{9C0006} -13.84\%} & {\color[HTML]{9C0006} -13.91\%} & {\color[HTML]{9C0006} -10.55\%} & {\color[HTML]{9C0006} -12.09\%} & {\color[HTML]{9C0006} -14.31\%} & {\color[HTML]{9C0006} -4.44\%}  \\
                                       & -                                     & 65.68\%                         & 57.28\%                         & 76.36\%                              & -                                    & -                                     & 99.12\%                         & \textbf{99.71\%}                & 99.01\%                         & 99.53\%                         & 98.80\%                         & 89.84\%                         \\
\multirow{-2}{*}{Number}               & {\color[HTML]{4EA72E} -}              & {\color[HTML]{4EA72E} +0.07\%}  & {\color[HTML]{9C0006} -2.41\%}  & {\color[HTML]{9C0006} -2.62\%}       & {\color[HTML]{4EA72E} -}             & {\color[HTML]{4EA72E} -}              & {\color[HTML]{9C0006} -0.03\%}  & {\color[HTML]{9C0006} -0.01\%}  & {\color[HTML]{9C0006} -0.02\%}  & {\color[HTML]{9C0006} -0.01\%}  & {\color[HTML]{9C0006} -0.12\%}  & {\color[HTML]{4EA72E} +0.60\%}  \\
                                       & -                                     & 68.18\%                         & 58.30\%                         & 70.66\%                              & -                                    & -                                     & 99.79\%                         & 99.93\%                         & 99.81\%                         & \textbf{99.98\%}                & 98.79\%                         & 92.43\%                         \\
\multirow{-2}{*}{Add Paragraph}        & {\color[HTML]{4EA72E} -}              & {\color[HTML]{4EA72E} +2.56\%}  & {\color[HTML]{9C0006} -1.40\%}  & {\color[HTML]{9C0006} -8.32\%}       & {\color[HTML]{4EA72E} -}             & {\color[HTML]{4EA72E} -}              & {\color[HTML]{4EA72E} +0.64\%}  & {\color[HTML]{4EA72E} +0.22\%}  & {\color[HTML]{4EA72E} +0.78\%}  & {\color[HTML]{4EA72E} +0.44\%}  & {\color[HTML]{9C0006} -0.13\%}  & {\color[HTML]{4EA72E} +3.18\%}  \\
                                       & {\ul 21.30\%}                         & 44.83\%                         & 20.32\%                         & 36.10\%                              & {\ul 66.20\%}                        & {\ul 9.30\%}                          & 89.59\%                         & \textbf{97.48\%}                & 66.89\%                         & 83.62\%                         & 81.20\%                         & 51.47\%                         \\
\multirow{-2}{*}{Homoglyph}            & {\color[HTML]{9C0006} {\ul -35.40\%}} & {\color[HTML]{9C0006} -20.78\%} & {\color[HTML]{9C0006} -39.38\%} & {\color[HTML]{9C0006} -42.88\%}      & {\color[HTML]{9C0006} {\ul -0.30\%}} & {\color[HTML]{9C0006} {\ul -75.70\%}} & {\color[HTML]{9C0006} -9.56\%}  & {\color[HTML]{9C0006} -2.23\%}  & {\color[HTML]{9C0006} -32.14\%} & {\color[HTML]{9C0006} -15.92\%} & {\color[HTML]{9C0006} -17.72\%} & {\color[HTML]{9C0006} -37.78\%} \\
                                       & {\ul 33.20\%}                         & 63.00\%                         & 48.94\%                         & 73.27\%                              & {\ul 61.00\%}                        & {\ul 71.40\%}                         & 98.60\%                         & 99.32\%                         & 98.02\%                         & \textbf{99.42\%}                & 98.52\%                         & 84.52\%                         \\
\multirow{-2}{*}{Article Deletion}     & {\color[HTML]{9C0006} {\ul -23.50\%}} & {\color[HTML]{9C0006} -2.61\%}  & {\color[HTML]{9C0006} -10.75\%} & {\color[HTML]{9C0006} -5.71\%}       & {\color[HTML]{9C0006} {\ul -5.50\%}} & {\color[HTML]{9C0006} {\ul -13.60\%}} & {\color[HTML]{9C0006} -0.55\%}  & {\color[HTML]{9C0006} -0.39\%}  & {\color[HTML]{9C0006} -1.01\%}  & {\color[HTML]{9C0006} -0.13\%}  & {\color[HTML]{9C0006} -0.40\%}  & {\color[HTML]{9C0006} -4.72\%}  \\
                                       & -                                     & 65.50\%                         & 58.23\%                         & 77.56\%                              & -                                    & -                                     & 99.02\%                         & \textbf{99.64\%}                & 98.94\%                         & 99.49\%                         & 98.80\%                         & 90.46\%                         \\
\multirow{-2}{*}{Alternative Spelling} & {\color[HTML]{4EA72E} -}              & {\color[HTML]{9C0006} -0.12\%}  & {\color[HTML]{9C0006} -1.46\%}  & {\color[HTML]{9C0006} -1.42\%}       & {\color[HTML]{4EA72E} -}             & {\color[HTML]{4EA72E} -}              & {\color[HTML]{9C0006} -0.13\%}  & {\color[HTML]{9C0006} -0.07\%}  & {\color[HTML]{9C0006} -0.09\%}  & {\color[HTML]{9C0006} -0.05\%}  & {\color[HTML]{9C0006} -0.11\%}  & {\color[HTML]{4EA72E} +1.21\%}  \\
                                       & -                                     & 78.40\%                         & 97.88\%                         & 98.41\%                              & -                                    & -                                     & \textbf{100.00\%}               & 98.63\%                         & 91.95\%                         & 72.81\%                         & 95.43\%                         & 68.00\%                         \\
\multirow{-2}{*}{Zero-Width Space}     & {\color[HTML]{4EA72E} -}              & {\color[HTML]{4EA72E} +12.79\%} & {\color[HTML]{4EA72E} +38.18\%} & {\color[HTML]{4EA72E} +19.43\%}      & {\color[HTML]{4EA72E} -}             & {\color[HTML]{4EA72E} -}              & {\color[HTML]{4EA72E} +0.85\%}  & {\color[HTML]{9C0006} -1.08\%}  & {\color[HTML]{9C0006} -7.08\%}  & {\color[HTML]{9C0006} -26.73\%} & {\color[HTML]{9C0006} -3.49\%}  & {\color[HTML]{9C0006} -21.25\%} \\
\midrule
                                       & {\ul 47.73\%}                         & 63.75\%                         & 50.73\%                         & 70.32\%                              & {\ul 63.92\%}                        & {\ul 72.90\%}                         & 96.30\%                         & \textbf{97.79\%}                & 94.15\%                         & 94.34\%                         & 95.10\%                         & 83.98\%                         \\
\multirow{-2}{*}{Avg attacks}          & {\color[HTML]{9C0006} {\ul -8.97\%}}  & {\color[HTML]{9C0006} -1.86\%}  & {\color[HTML]{9C0006} -8.96\%}  & {\color[HTML]{9C0006} -8.66\%}       & {\color[HTML]{9C0006} {\ul -2.58\%}} & {\color[HTML]{9C0006} {\ul -12.10\%}} & {\color[HTML]{9C0006} -2.85\%}  & {\color[HTML]{9C0006} -1.92\%}  & {\color[HTML]{9C0006} -4.88\%}  & {\color[HTML]{9C0006} -5.20\%}  & {\color[HTML]{9C0006} -3.82\%}  & {\color[HTML]{9C0006} -5.27\%}  \\
\bottomrule
\end{tabular}
\end{adjustbox}

\caption{Recall at 5\% FPR on the RAID dataset by adversarial attack. Results in the first box are from models trained on datasets other than RAID, and are evaluated on the official test set. Results in the second box are pre-trained on MAGE and fine-tuned for one epoch on RAID. Results in the third box are fine-tuned for one epoch on RAID, but pre-trained on other datasets. Both of them are trained and tested on our own data split. Further, underlined results are taken directly from~\cite{dugan2024RAIDShareda}, as the detectors's predictions and results were unavailable on their repository. We show both the absolute value of the metric (top) and its difference with the base performance without adversarial attacks (bottom).}
\label{tab:raid-attack-results}
\end{table}

\section{Conclusions}\label{sec:conclusions}

In this work, we proposed the Perplexity Attention Weighted Network (PAWN) for AI-generated text detection. Zero-shot methods typically aggregate next-token distribution metrics using a simple mean. Based on the idea that some tokens are naturally easier (e.g. word completions) and harder (e.g. beginnings of texts without prompt conditioning) to predict, we use semantic information from the last hidden states and positional information to weight features based on next-token distribution metrics before aggregating them across the sequence length.

\paragraph{Strengths}
Although not a zero-shot method, our method employs a very small number of trainable parameters, in the order of one million. In addition, since the LLM is frozen, we can cache the hidden states and next-token distribution metrics of training texts on disk, greatly reducing the training resource requirements.

PAWN performs competitively or even better in-distribution than the best baselines, which are fine-tuned LMs. It also generalizes better overall to new models and domains, especially in terms of decision boundaries, as fine-tuned LMs tend to maintain a high AUROC out-of-distribution but drop recall performance due to variations in the optimal boundary.

PAWN shows very strong performance even with a small backbone such as \texttt{openai-community/\allowbreak gpt2}, but a medium backbone such as \texttt{meta-llama/\allowbreak Llama-3.2-1B-Instruct} is slightly stronger in general. It also has better multilingual capabilities, providing decent generalization in languages not seen during supervised training. Finally, PAWN models present a better average robustness to a diverse set of adversarial attacks compared with the selected baselines.

\paragraph{Weaknesses}
Despite the robustness of PAWN to many types of adversarial attacks, paraphrasing attacks are still challenging for our models and further work is still required. Training frameworks like RADAR might be applied in future work to improve robustness in this regard. 

While the frozen LLM backbone allows us to train very efficiently by caching metrics and hidden states, inference times are still very affected by the large size of the backbones models we used. This is specially true for our best performing backbone \texttt{meta-llama/\allowbreak Llama-3.2-1B-Instruct}, with 1 billion parameters, many more than traditional encoder-only models like RoBERTa.

\paragraph{Future ways} PAWN can be leveraged by others as a strong and more robust alternative to fine-tuned LMs in detecting AI-generated text. Some training frameworks that are applied to the latter, such as RADAR, can also be applied to PAWN, possibly pushing even further state-of-the-art performance. These types of improvements will be specially critical to improve PAWN's robustness to paraphrasing attacks, a major weakness that must be addressed before applying it in real-world scenarios.

While \cref{sec:qualitative analysis} provides a brief analysis of the patterns that PAWN is learning, further study is necessary to fully understand the model. Studying the performance of PAWN with a larger set of backbones might also lead to interesting discoveries. Finally, a deeper hyperparameter study in PAWN, especially the number of gates, might improve the model's interpretability without sacrificing detection performance.

\section*{Acknowledgments}
This work has been supported by the project PCI2022-134990-2 (MARTINI) of the CHISTERA IV Cofund 2021 program; by European Comission under IBERIFIER Plus - Iberian Digital Media Observatory (DIGITAL-2023-DEPLOY- 04-EDMO-HUBS 101158511), and by TUAI Project (HORIZON-MSCA-2023-DN-01-01, Proposal number: 101168344) projects; by EMIF managed by the Calouste Gulbenkian Foundation, in the project MuseAI; and by Comunidad Autonoma de Madrid, CIRMA-CM Project (TEC-2024/COM-404).

\printbibliography

\appendix
\section{PyTorch pseudocode of PAWN}\label{sec:pseudocode}

\begin{lstlisting}[
    caption={PyTorch pseudocode of PAWN},
    label={lst:pseudocode}
]
def forward(self, texts: list[str]) -> torch.Tensor:
    """
    Returns the logits of the model for the given texts.
    """
    hs, metrics, attention_mask = self.pretrained_model.forward(texts)  # [B, L, H], [B, L-1, M], [B, L]
    attention_mask = attention_mask[:, :-1]  # [B, L-1]
    B, L, H = hs.size()

    # Apply metrics_nn to obtain metrics features
    processed_metrics = self.metrics_nn(metrics)  # [B, L-1, F]
    
    # Create semantic and positional embeddings
    pos_embed = torch.arange(L-1, device=hs.device).unsqueeze(0).expand(B, -1)  # [B, L-1]
    pos_embed = pos_embed.unsqueeze(-1) / self.max_len  # [B, L-1, 1]
    gate_x_list = [hs[..., :-1, :], hs[..., 1:, :], pos_embed]
    gate_x = torch.cat(gate_x_list, dim=-1)

    # Dropout a percentage of tokens for regularization
    gate_mask = self.__dropout_tokens(attention_mask == 0)  # [B, L-1]
    
    # Apply gate_nn to obtain gate logits, and make sure to mask padding and dropped tokens
    gate_logits = self.gate_nn(gate_x)  # [B, L-1, G], where G divides F
    gate_logits = gate_logits.masked_fill(gate_mask.unsqueeze(-1), float("-inf"))  # [B, L-1, G]

    # If the number of gates is less than the number of features, repeat the gate logits
    G, F = gate_logits.size(-1), processed_metrics.size(-1)
    if 1 < G < F: gate_logits = gate_logits.repeat(1, 1, F // G)  # [B, L-1, F]

    # Apply the softmax across seq len to obtain weights
    gates = gate_logits.softmax(dim=-2)  # [B, L-1, F]
    # Apply the attention mechanism to obtain the weighted sum of the features
    aggregate_output = gates * processed_metrics  # [B, L-1, F]
    aggregate_output = aggregate_output.sum(dim=-2)  # [B, F]
    
    # Apply the final MLP to obtain the logits
    logits = self.aggregate_nn(aggregate_output)  # [B, 1]
    return logits.squeeze(-1)
\end{lstlisting}

\section{Qualitative analysis of weights in PAWN}\label{sec:qualitative analysis}

In this section we try to illustrate and investigate PAWN by studying the weights that it assigns to different tokens. To do this, we take a checkpoint trained on MAGE with the LLaMA-1b backbone. We first show the effect of the token's position on the attention they receive, and then some of the highest and lowest weighted tokens on average.

The reader will observe that each token doesn't receive a single weight. To be more specific, the position and hidden states produce \emph{weight logits} in \(\mathbb R ^{L \times G}\), where \(L\) is the number of tokens or sequence length and \(G\) is the number of gates. In this particular checkpoint we have \(G = 256\). To produce the weights, we apply the softmax operation across the first dimension or sequence length. This yields \(256\) different weights for each token. We attempt to make this more interpretable by considering the average of these weights as a single number, but we warn that the volatility is high across the different weights, and that this is a very big simplification.

\subsection{Average weights by position}

\begin{figure}[t]
    \centering
    \includegraphics[width=0.7\linewidth]{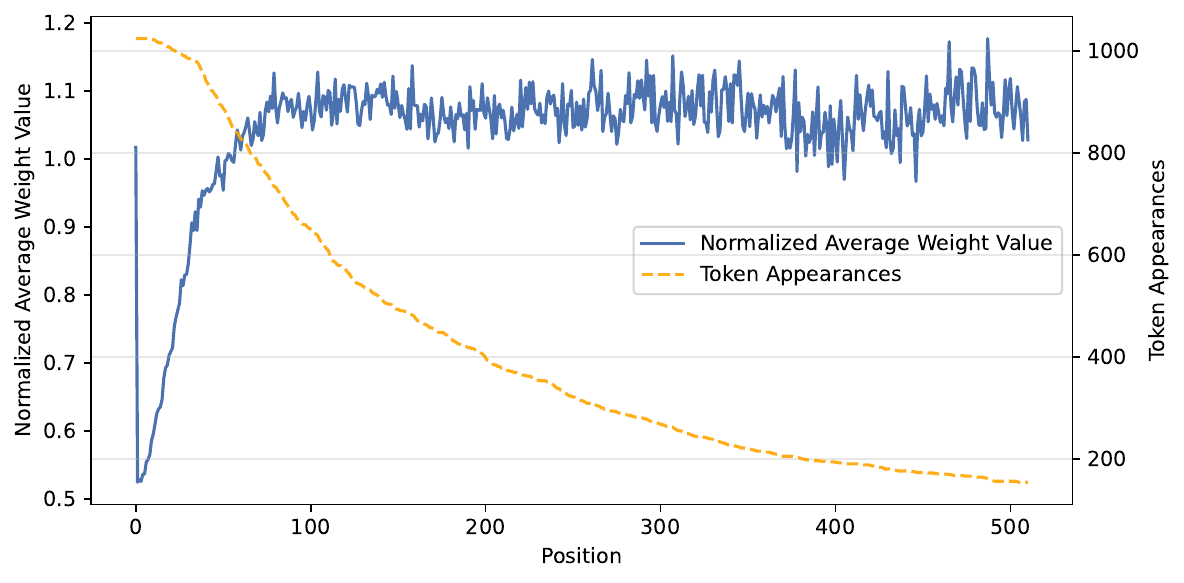}
    \caption{Average weights given to each token position across 1024 samples, normalized by multiplying with the number of tokens in the sample. The number of tokens that appear in that position is also shown.}
    \label{fig:weights-by-pos}
\end{figure}

In \cref{fig:weights-by-pos} we show the average weights given to each token position across 1024 samples. Since weights sum up to one, they are much smaller for longer sequences. Thus, we normalize by multiplying the attention weights of each sample with the number of tokens. As we can see, the average weight assigned to each token increases monotonically at a good rate up to around token 100. After this, the weights become stable and uniform. There is one notable exception: the very first token seems to receive a lot of attention. This might be due to biases of modern LLMs in the way they start texts, but we have no certainty here.

\subsection{Average weights by token}

In \Cref{tab:weights-by-token} we show the highest and lowest weighted tokens on average from a sample of 1024 examples, filtering out tokens with less than 10 appearances. As before, we normalize the weights by the number of tokens in the example. Regarding the tokens with the lowest average weight, we see some groups of numbers, some word terminations (e.g. ``ington'' or ``chlor'') and some individual characters. In the highest weighted token, we observe some well-known biases of LLMs: starting the sentence with ``Additionally'', ``Furthermore'' or ``Overall''.

\begin{table}[t]
    \centering
    \begin{adjustbox}{max width=0.7\textwidth}
    \begin{tabular}{cc|cc}
\toprule
Token        & Avg. Norm. Weight             & Token  & Avg. Norm. Weight             \\
\midrule
Additionally & 2.5112                        & San    & 0.2731                        \\
prediction   & 2.4350                        & 197    & 0.2886                        \\
bacteria     & 2.1545                        & 202    & 0.2917                        \\
message      & 2.1354                        & 196    & 0.3143                        \\
addition     & 2.1106                        & ton    & 0.3238                        \\
activities   & 2.0291                        & Ch     & 0.3256                        \\
Furthermore  & 2.0233                        & 201    & 0.3259                        \\
directions   & 1.9634                        & O      & 0.3271                        \\
solutions    & 1.9106                        & \{     & 0.3286                        \\
steps        & 1.9039                        & 199    & 0.3321                        \\
playlist     & 1.8674                        & chlor  & 0.3525                        \\
undercut     & 1.8560                        & R      & 0.3527                        \\
Overall      & 1.8466                        & 198    & 0.3535                        \\
According    & 1.8462                        & il     & 0.3543                        \\
protests     & 1.8383                        & New    & 0.3661                        \\
algorithms   & 1.8308                        & e      & 0.3753                        \\
authorities  & 1.8236                        & w      & 0.3850                        \\
alone        & 1.8187                        & ul     & 0.3895                        \\
instance     & 1.8157                        & ington & 0.3901                        \\
algorithm    & 1.8005                        & CM     & 0.3903                        \\
\bottomrule
\end{tabular}
    \end{adjustbox}
    \caption{Highest (left) and lowest (right) weighted tokens on average from a sample of 1024 examples, filtering out tokens with less than 10 appearances.}
    \label{tab:weights-by-token}
\end{table}

\end{document}